\RequirePackage{snapshot}
\documentclass[doublespacing]{elsart}

\usepackage{ucs}
\usepackage[utf8x]{inputenc}
\usepackage{url}
\usepackage[cmex10]{amsmath}
\usepackage{amssymb}
\usepackage{color}
\usepackage{graphicx}
\usepackage{subfigure}
\usepackage{pgfplots}

\newcommand{\makeinterval}[2]{ $ [ #1, #2 ] $ }
\newcommand{\degree}{\ensuremath{^\circ}}
\newcommand{\mytilde}{\raise.17ex\hbox{$\scriptstyle\mathtt{\sim}$}}

\newcommand{\DONE}[1]{}

\graphicspath{{figures/}}

\begin{document}

\begin{frontmatter}



\title{DSeg: Direct Line Segments Detection}


\author{Cyrille Berger\corauthref{cor1}}
\ead{cyrille.berger@liu.se}
\ead[url]{http://www.ida.liu.se/\mytilde cyrbe}
\author{Simon Lacroix}
\ead{simon.lacroix@laas.fr}
\ead[url]{http://homepages.laas.fr/simon}

\address{CNRS, LAAS, 7 avenue du colonel Roche, F-31400 Toulouse, France\\Univ de Toulouse, LAAS, F-31400 Toulouse, France}

\corauth[cor1]{Corresponding author}

\begin{abstract}

  This paper presents a model-driven approach to detect image line segments. The
  approach incrementally detects segments on the gradient image using a linear
  Kalman filter that estimates the supporting line parameters and their
  associated variances. The algorithm is fast and robust with respect to image
  noise and illumination variations, it allows the detection of longer line
  segments than data-driven approaches, and does not require any tedious
  parameters tuning. An extension of the algorithm that exploits a pyramidal
  approach to enhance the quality of results is proposed. Results with varying
  scene illumination and comparisons to classic existing approaches are
  presented.
\end{abstract}


\end{frontmatter}

\section{Introduction}
%
%
%
%

\begin{figure}[b!]
\centering
\includegraphics[width=\linewidth,trim=1cm 4.5cm 5cm 1cm]{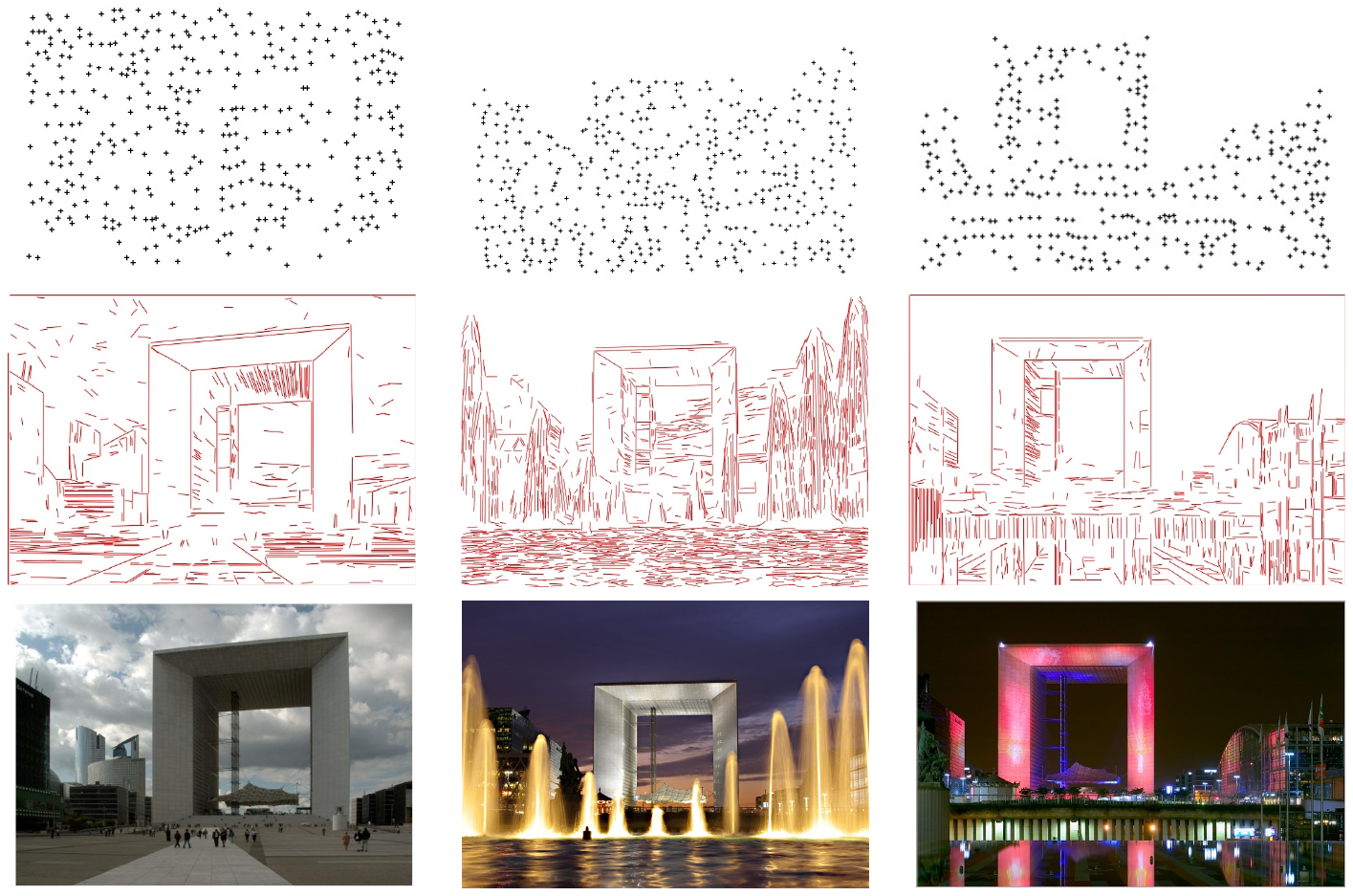}
\caption{Harris feature points and line segments detected on three images of
  the same scene, under large illumination variations.}
\label{fig:segvpts}
\end{figure}

Image line segments carry a lot of information on the perceived scene structure,
and are basic primitives on which several vision applications are built on ({\em
  e.g.} structured object recognition \cite{DAVID-2005}, or structured scenes 3D
reconstruction in geomatics or robotics
\cite{SMITH-2006,SOLA-IROS-2009}). Figure \ref{fig:segvpts} illustrates that
contrary to interest points, line segments can capture the gist of a scene
regardless of large illumination variations.

Line segments have numerous advantageous properties: they of course perfectly
represent the straight lines of a scene, they can be used to estimate geometric
transformations such as homographies or 3D transformations, and they are
invariant to large viewpoint and scale changes. But this latter property
actually only holds if the line segment extraction process is itself {\em
  robust} with respect to image noise. A stable, robust and fast segment
detection 
is therefore desirable, and is still the object of recent contributions
\cite{CHEN-2007,GROMPONE-2008}.


\paragraph*{Related work.}


The first classical approach to the line segment detection problem relies on the
Canny filter \cite{CANNY-1986}, a four steps algorithm, which starts by smoothing
the image and computing the magnitude and gradient direction, then edges are thinned
and thresholded. This provide an edge-map that can be used in a \textit{chaining}\cite{ETEMADI-1992}
line fitting process to obtain the contours that define the line segments.


An other classical approach also exploits the Canny filter, and computes a Hough
transform on the edge map \cite{HOUGH-1962} where each Canny points votes for
all line parameters passing through that point. Since this method only allows finding
the infinite line supports, the application of a splitting algorithm is necessary to find the
line segments. Many improvements to this approach have been introduced over the
years: in \cite{BALLARD-PR-1981}, Ballard proposed the extension of the voting procedure to
use the gradient direction information, random samples of the observations are
used to speed up the process in \cite{KIRYATI-1991}, the votes are weighted to
give more importance to pixels with a clear boundary in \cite{GUO-2008}, a
Kalman filter is used to improve the voting procedure of the Hough transform in
\cite{XU-1994}, and a hierarchical approach is proposed in \cite{YACOUB-1995}.


Other approaches rely on detecting regions of pixels with similar gradient
orien\-ta\-tion, to which line segment parameters are fit afterward. In
\cite{BURNS-1986}, pixels with similar gradient orientations are grouped, and
then line parameters are estimated from the rectangle they define. In
\cite{DESOLNEUX-2000}, Desolneux proposes a statistical framework to assess
whether a sufficient number of pixels in a region are aligned on the basis of
their gradient. Recently, Grompone proposed a Line Segment Detector (LSD)
\cite{GROMPONE-2008} built upon the work of Burns \cite{BURNS-1986} and
Desolneux \cite{DESOLNEUX-2000}, mixing both approaches and offering the
following improvements:

%
%

\begin{enumerate}
\item As in \cite{BURNS-1986}, the image is first segmented with a region
  growing algorithm using the gradient direction as a grouping criteria.
\item Regions are then approximated as a rectangle, using the center of mass of
  the region as the center of the rectangle \cite{KAHN-1990}, the gradient value
  to define the weight of each pixel, and the first inertia axis to select the
  orientation.
\item The last step applies the idea of \cite{DESOLNEUX-2000} to the found
  rectangle, by adjusting the parameters of the rectangle until a linear segment
  is found.
\end{enumerate}
One of the main improvement of this method over the previous ones is that it
does not require any tuning of parameters -- it also gives more accurate
segments.

\DONE{Verifier si DESOLNEUX-2000 In section 2.1: the algorithm in [13] does not
rely on detecting region and then fitting line segment parameters: actually in
that article they will only detect pixel alignment, but they don't consider any
fitting of the segment, but if they have two lines of pixels with similar
gradient, they will detect both lines, it is separated in GROMPONE-2008}

\paragraph*{Overview of the proposed approach.}

This article presents an approach to line segment detection that is directly
fits segment models on the data, using measures on the gradient to adjust their
parameters\footnote{the work depicted here provides a more in-depth description
  of an approach previously published in \cite{BERGER-RFIA-2010}, as well as
  quantitative comparisons and a hierarchical extension.}. The approach relies on
a seed-and-grow scheme and a Kalman filter: null length line segments are
initialized on local gradient maxima, and further extended and updated thanks to
the Kalman filter that estimates the line segment parameters by incrementally
augmenting the line length. It is the current line parameters that drive the
search for additional supporting gradients in the data, instead of the
neighboring gradients that drive the update of the line parameters. Similar
approaches have been proposed: \cite{CHEN-2007} proposes to introduce some
knowledge of the model in the grouping step, and \cite{MANSOURI-1987} suggests
generating a segment hypothesis for each pixel based on the gradient
orientation, and then using a statistic approach to confirm the
hypothesis. Compared to \cite{MANSOURI-1987}, the originality of the proposed
method lies in the use of a Kalman filter to estimate the parameters of the
segment model, hence avoiding the need for thresholding the gradients or the use
of a heuristic split/merge process.

%

The Kalman filter is commonly used in tracking and especially for tracking segments
between consecutive frames in a video sequence \cite{MILLS-BMVC-2003} and for computing
the 3D geometric parameters of the segment \cite{SMITH-2006}. The originality of our
method comes from the use of the Kalman filter to estimate the segment parameters
during the detection step.

The main interests of our approach is that it is fast and it extracts longer lines
than the classical approaches.
It is robust with respect to image noise and scene illumination
variations. Most importantly, it does not require the tuning of
parameters, an issue which most existing methods have, their
results being very sensitive to the choice of thresholds -- with the notable
exception of \cite{GROMPONE-2008} and \cite{MANSOURI-1987}.

We also provide an open-source implementation of our algorithm in ~\cite{libsegments}.


\paragraph*{Outline.} The next section is the core of the article: it depicts
\textit{DSeg}, the proposed line segments detection algorithm based on a Kalman
filter A hierarchical version of the algorithm is proposed in section
\ref{sec|hdseg}. A comparison with four other detection schemes is made in
section \ref{sec|comparision}, in terms of number and length of detected
segments, sensitivity with respect to noise and illumination
variations. 



\section{Detection}
\label{sec|dseg} 

\subsection{Overview}

\begin{figure}[tb]
  \centerline{
    \includegraphics[width=0.24\linewidth]{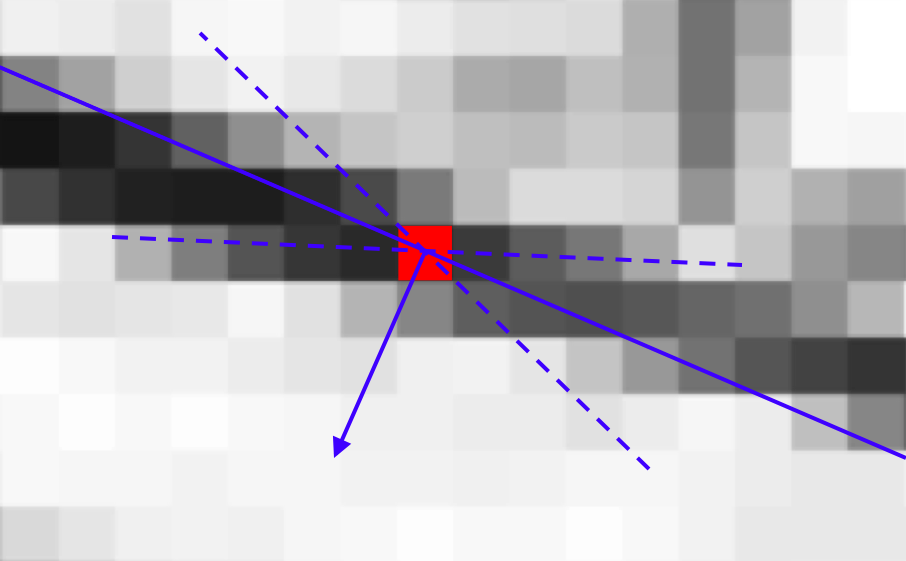}
    \includegraphics[width=0.24\linewidth]{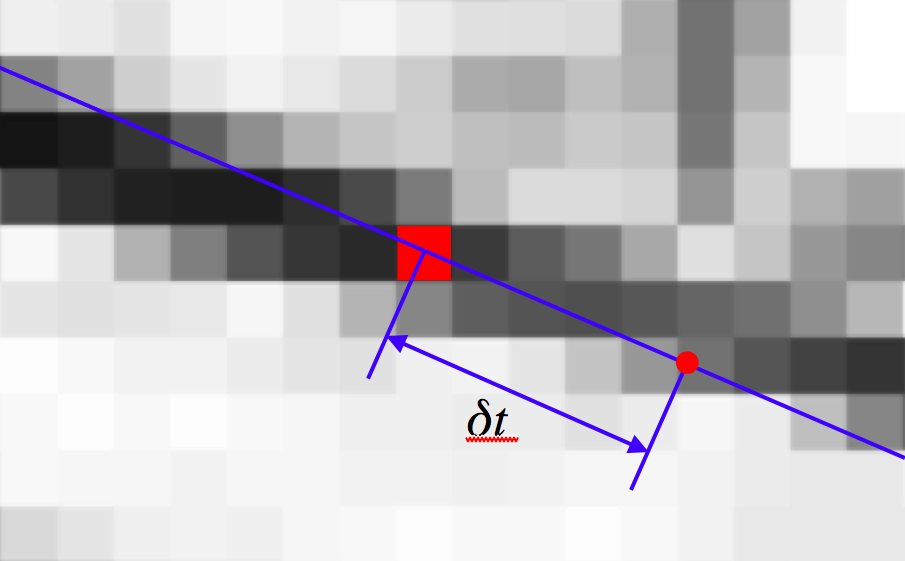}
    \includegraphics[width=0.24\linewidth]{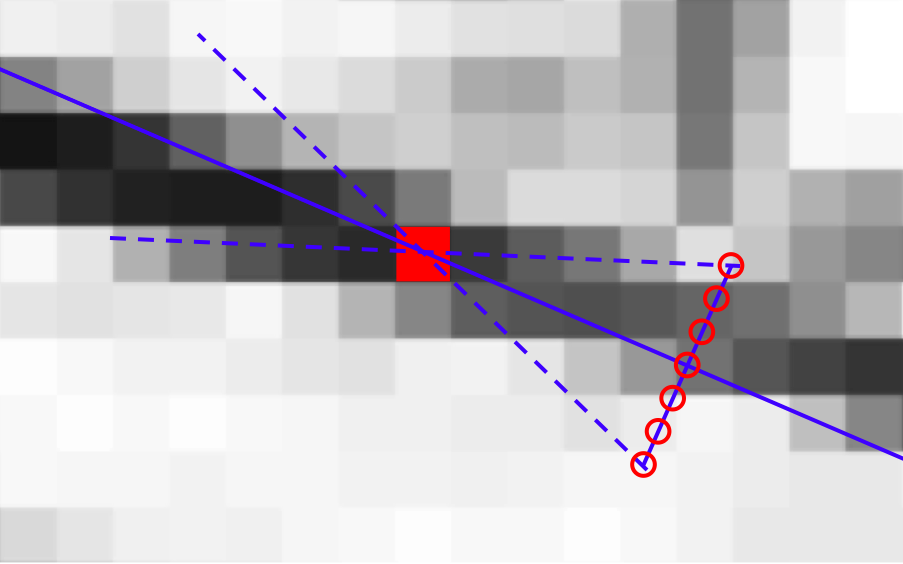}
    \includegraphics[width=0.24\linewidth]{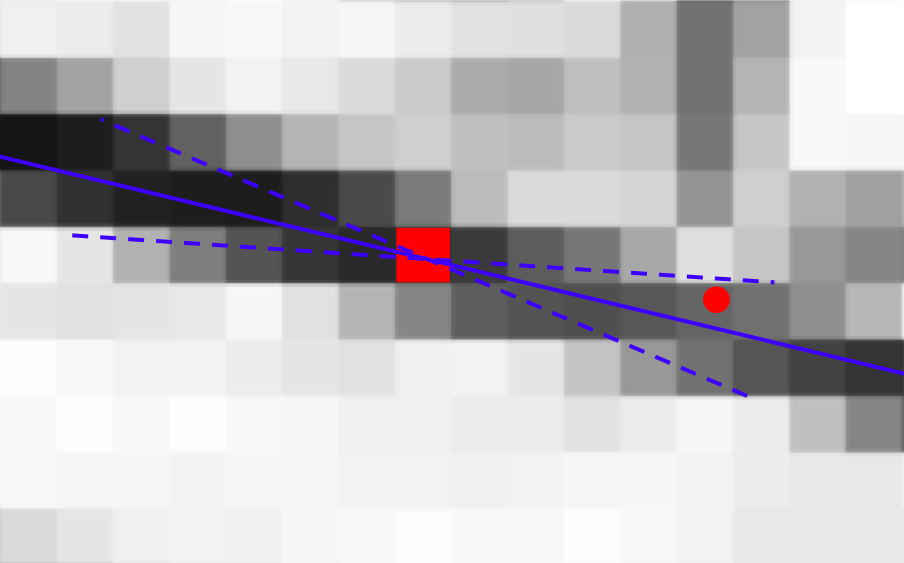}
  }
  \centerline{
    \parbox{0.24\linewidth}{\center (a) }
    \hfill
    \parbox{0.24\linewidth}{\center (b) }
    \hfill
    \parbox{0.24\linewidth}{\center (c) }
    \hfill
    \parbox{0.24\linewidth}{\center (d) }
}
\caption[]{Illustration of the steps of the Kalman filter to detect line
  segments on a gradient image. (a) initialization: a line segment initialized
  on the basis of the local maximum gradient (square red pixel, first support
  point of the line segment): the arrow indicates the gradient orientation, and
  the dashed lines represent the initial uncertainty on the direction of the
  associated supporting line. (b) prediction: an additional support point (red
  point) is predicted at a distance $\delta t$ from the last support point --
  $\delta t$ is here equal to several pixels for readability reasons. (c)
  observation: a set of gradient measures (red circles) defined across the
  prediction is analyzed. (d) update: if a valid observation is found (red
  point), the supporting line parameters and associated variances are updated.}
  \label{fig|KalmanSteps}
\end{figure}

The detection process fits a line segment model on the image using the image
gradients, thanks to a Kalman filter scheme. Its steps are illustrated Figure
\ref{fig|KalmanSteps}:
 
\begin{itemize}
\item \textbf{Initialization:} a line segment is initialized on the sole basis of the
  gradients values and orientations computed on the image
  (Fig. \ref{fig|KalmanSteps}-a). It would be naturally very time consuming to
  search for segments around every pixel: the process is therefore only
  initialized for {\em seeds}, {\em i.e.}  pixels that correspond to a local
  gradient maximum, as in \cite{MANSOURI-1987}.

  Once the line segment is initialized, additional support points along the
  current estimated direction according to the Kalman filter predict /
  observe / update sequence.

\item \textbf{Prediction:} on the basis of the current estimated supporting line, the
  next support point location is estimated at a given distance $\delta t$
  (Fig. \ref{fig|KalmanSteps}-b);

\item \textbf{Observation:} a set of gradients measures along the normal to the predicted
  support point are analyzed, within ranges set by the current uncertainty on
  the support line orientation (Fig. \ref{fig|KalmanSteps}-c). Among these
  measures, a gradient maximum that satisfies some conditions is declared a
  valid observation of the current line segment.
  
\item \textbf{Update:} on the basis of the prediction and the observation, the 
  Kalman filter update is applied, resulting in a new estimate of the support
  line parameters and of its associated variances
  (Fig. \ref{fig|KalmanSteps}-d).

\end{itemize}

The steps prediction, observation and update are applied alternating each side of the segment.

The linear Kalman formulation was used rather than a linear least-square since the linear Kalman filter allows to efficiently compute predictions for observations and incremental updates while being equivalent to a least-square.

This process is iterated to extend the line segment as long as valid
observations are found.

\subsection{Setting up the Kalman filter}

\subsubsection{System equations}

A line is represented {\em in the image frame} with the following parametric linear
model:
\begin{eqnarray}
 x(t) = at + x_0 \\
 y(t) = bt + y_0
\end{eqnarray}

where $ (a, b) $ is the direction vector of the line and $ (x_0, y_0 ) $ its
origin. Although this model is over-parametrized, it brings forth two essential
advantages: it can represent any 2D line (no singularity), and the single
parameter $t$ allows specifying any point on the line. The fact that this model
is not minimal is not an issue, because no constraints link its parameters.

The state in the Kalman filter is the set of line parameters: $ \mathbf{x_{k|k}} =
(a,x_0,b,y_0)^T $ (bold notation denotes vectors) of covariance $ P_{k|k} $. Since
there is no process noise and the model is stationary, the state equation is:
\begin{align}
\mathbf{x_{k|k-1}} & = \mathbf{x_{k-1|k-1}} \\
 P_{k|k-1} & = P_{k|k}
\end{align}

The observation equation is:
\begin{equation}
\mathbf{z_k} = H_k \mathbf{x_k} + \mathbf{v_k}
\end{equation}

It is a linear observation model, where $v_k$ is the observation noise with
covariance $R = \sigma_r^2 I_2$, and $H_k$ the observation matrix:
\begin{align}
H_k = &
\begin{bmatrix}
 (\gamma \cdot t) & 1 & 0 & 0 \\
 0 & 0 & (\gamma \cdot t) & 1
\end{bmatrix} \\
t = & k \cdot \delta_t
\end{align}

$\gamma = 1$ or $\gamma = -1$ depending on the extension direction.

\subsubsection{Filter initialization}

A line segment is initialized on local maxima of gradient values. Let $G_{i,j}$
and $\phi_{i,j}$ the norm and the direction of the gradient on pixel $(i,j)$,
computed using $3\times3$ Sobel kernels.  A pixel $(i,j)$ is considered as a
seed to initialize a filter if it satisfies the following conditions:

\begin{eqnarray}
  G_{i,j} - G_{i + c_\phi(i,j), j + s_\phi(i,j)} > \tau_{Gmax} \\
  G_{i,j} - G_{i - c_\phi(i,j), j - s_\phi(i,j)} > \tau_{Gmax} \\
  G_{i,j} > G_{i - s_\phi(i,j), j + c_\phi(i,j)} \\
  G_{i,j} > G_{i + s_\phi(i,j), j - c_\phi(i,j)}
\end{eqnarray}

where $c_\phi(i,j) = cos(\phi_{i,j})$ and $s_\phi(i,j) = sin(\phi_{i,j})$ (in
all the algorithm steps, subpixel values are obtained by bi-cubic
interpolations, applied on the gradient values for computation efficiency
reasons). The first two conditions ensure that the considered pixel is likely to
belong to a line segment, while the two others state that the considered pixel
gradient is a local maximum along the hypothetical line segment: they are
mostly introduced to give more robust seeds.

Given a seed, the parameters of the line model that initialize the state of the
Kalman filter are:

\begin{eqnarray}
  x_0 = i\\
  y_0 = j\\
  a = -s_\phi(i,j)\\
  b = c_\phi(i,j)
\label{equ|init}
\end{eqnarray}

and the associated covariances are initialized to (see section
\ref{sec|parameters}):

\begin{equation}
P_{0|0} =
\begin{bmatrix}
 \sigma_{a}^2 & 0 & 0 & 0 \\
 0 & \sigma_{x0}^2 & 0 & 0 \\
 0 & 0 & \sigma_{b}^2 & 0 \\
 0 & 0 & 0 & \sigma_{y0}^2 \\
\end{bmatrix}
\end{equation}

\subsubsection{Prediction}

The process extends a line segment and refines the supporting line parameters by
searching for additional support points in the gradient image. Given a current
support point, a new support point $(x_{t+\delta t}, y_{t+\delta t})$ is simply
defined by applying the state equation to the value $t+\delta t$. The scalar
error $e_{t+\delta t}$ represents the error across the current estimated line
at the predicted support point, it is computed from the innovation covariance~\cite{BAR-EATN-2004}:

\begin{equation}
S_k = H_{k+1} P_{k|k} H_{k+1} + \sigma_r^2 I_2
\end{equation}

Where $\sigma_r$ is the standard deviation of the observation noise.
Lets call $\sigma_p $ the largest eigen value of $S_k$:

\begin{equation}
e_{t+\gamma\delta t} = 3 \sigma_p
\end{equation}

Note that right after the initialization ($t=0$), the support point is the
selected seed, and also that a line segment is extended in both directions.

\subsubsection{Observation}
\label{sssec|observation} 

Additional support points are searched along a direction normal to the current
line, within a range cross-distance to the line that depends on the variance of
the current segment estimation. The ones that correspond to a local gradient
maximum along its direction are considered as {\em observations} of the current line segment. When
there is more than one gradient maximum among the measure points, the closest to
the current line is selected as an observation, and when two local maximum are
at an equal distance to the line, the one with the direction closest to the
current line orientation is selected. This observation process is depicted in
details here.

A set $M$ of $2n_o+1$ gradient measures is defined along the normal of the
predicted support point as follows:

\begin{itemize}
\item the direction vector $\mathbf{n}$ of the normal to the current line segment
  is computed as: $\mathbf{n} = (a',b')^T$, where $a' = a / \sqrt{a^2 + b^2}$
  and $b' = b / \sqrt{a^2 + b^2}$. 
\item the set $M$ of measures is defined as 
\begin{equation}
  M = \{ (x_{t+\gamma \delta t}, y_{t + \gamma \delta_t}) + l \mathbf{s} \}, l \in [-n_o, n_o]
\end{equation}
where $\mathbf{s} = (e_{t+\gamma \delta t} / n_o)\mathbf{n}$ is a vector of length
$(e_{t+\gamma \delta t} / n_o)$ that defines the sampling of the measures along the
normal.

\end{itemize}

The selection process defines the measures that will be incorporated as a
support point (an observation) to update the line model. A measure $ m_l \in M $
can be selected as an observation if it satisfies the two conditions:

\begin{itemize}
\item The gradient must be a local maximum:
  \begin{eqnarray}
    G\big( (x_{t + \delta t}, y_{t + \gamma \delta t}) + l \mathbf{s} \big) > G\big( (x_{t + \gamma \delta t}, y_{t + \gamma \delta t}) + (l+1) \mathbf{s} \big) \\
    G\big( (x_{t + \delta t}, y_{t + \gamma \delta t}) + l \mathbf{s} \big) > G\big( (x_{t + \gamma \delta t}, y_{t + \gamma \delta t}) + (l-1) \mathbf{s} \big)
  \end{eqnarray}
\item The gradient direction is compatible with the current estimated line
  segment\footnote{the cosine is used to avoid problems caused by angle periodicity}:
  \begin{equation}
    cos\Big( \phi\big( (x_{t + \gamma \delta t}, y_{t + \gamma \delta t}) + l \mathbf{s} \big) - arctan(a/-b) \Big) >
    \tau_{angle}
  \end{equation}
\end{itemize}

If two measures $m_{l1}$ and $m_{l2}$ pass these tests, their distance to the
line is first checked, and in case of equal distances, the measure with the most
compatible gradient direction with the current line is selected: {\em e.g.}
$m_{l1}$ is selected if $|l_1| < |l_2|$, and if $|l_1| = |l_2| $, $m_{l1}$ is selected if:
\begin{equation}
  cos( \phi_{ (x_{t + \delta t}, y_{t + \delta t}) + l_1 \mathbf{s} } - arctan(a/-b) )
  > cos( \phi_{(x_{t + \delta t}, y_{t + \delta t}) + l_2 \mathbf{s} } - arctan(a/-b) )
\end{equation}

When no further support points are found, the search is extended a step
$\delta_t$ further ahead to avoid spurious segment splits due to image noise.
The extension process ends when when no compatible pixel is found in two
consecutive attempts.

%
%

The observation error is defined by two uncorrelated variances in the frame
associated with the current line estimate, that are set by the discretization
applied for the line extension process. The along-track variance is
$\delta_t^2$, the cross-track variance is $\sigma_r^2$, error of the image
discretization -- note that these are conservative values. In the frame
associated to the line, the observation error matrix is:

\begin{equation}
\begin{bmatrix}
 \delta_t^2 & 0 \\
 0 & \sigma_r^2
\end{bmatrix}
\end{equation}

A frame transformation is applied to obtain the observation error matrix $R_k$ in
the image reference frame:

\begin{equation}
R_k = R 
\begin{bmatrix}
 \delta_t^2 & 0 \\
 0 & \sigma_r^2
\end{bmatrix}
R^T
\end{equation}

where $R = \begin{bmatrix}
  cos(\alpha) & -sin(\alpha) \\
  sin(\alpha) & cos(\alpha)
\end{bmatrix}$ is the rotation matrix between the line frame and the image
reference frame, $\alpha$ being the current estimated angle of the line in the
image reference frame.

\begin{equation}
 \alpha = arctan\left(\frac{b}{a}\right)
\end{equation}

\DONE{Relate $\alpha$ to $a$ and $b$}

This defines all the parameters required for the update of the line segment
parameters with the linear Kalman filter when a new observation (support point)
is assessed.

\subsubsection{Update}

The classical Kalman filter update is used~\cite{BAR-EATN-2004}:

\begin{align}
  K_k              & = P_{k|k-1}H_{k}^T (H_k P_{k|k-1} H_k^T + R_k)^{-1} \\
  \mathbf{x_{k|k}} & = \mathbf{x_{k|k-1}} + K_k (\mathbf{z}_k - H_k \mathbf{x_{k|k-1}} ) \\
  P_{k|k}          & = (I - K_k H_k) P_{k|k-1}
\end{align}

%
%

\subsection{Line merging}

%
%

\begin{figure}[htb!]
  \centering
  \includegraphics[width=0.48\linewidth]{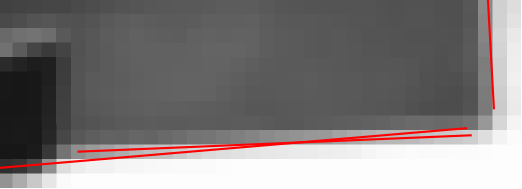}
  \hfill
  \includegraphics[width=0.48\linewidth]{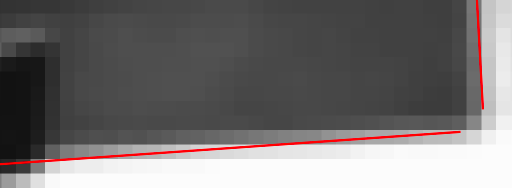}
  \caption[]{Line merging process. Left, two overlapping segments. Right, the line resulting from the merging
process.}
  \label{fig|closeUpMergeSegment}
\end{figure}

After the detection process, it can occur that a newly detected line segment
$S_n$ overlaps with a previously detected segment $S_p$ (figure
\ref{fig|closeUpMergeSegment}). In such cases, a merging process is applied: a
$\chi^2$ test is applied to check whether the extremities of $S_p$ can be
considered as observations (support points) compatible with $S_n$. The threshold
used for the $\chi^2$ test is the observation error for the point projection on
$S_n$. If yes, $S_n$ and $S_p$ are merged, and the four parameters of $S_n$ (as
well as their variances) are fused with the ones of $S_p$ using a weighted
average of Gaussian variables.

\begin{figure}[tb]
  \centering
  \includegraphics[width=0.45\linewidth]{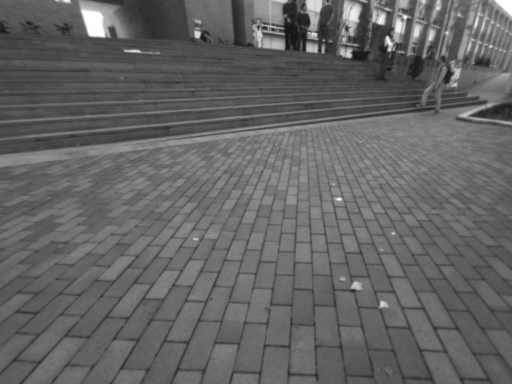}
  \includegraphics[width=0.45\linewidth, trim=0 8.8mm 0 0]{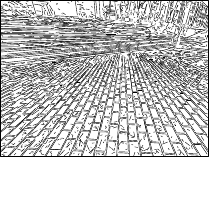}
  \includegraphics[width=0.45\linewidth]{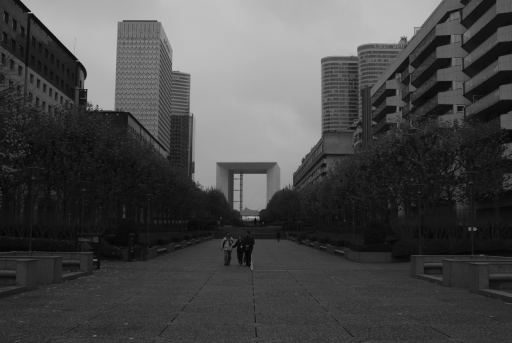}
  \includegraphics[width=0.45\linewidth, trim=0 11.6mm 0 0]{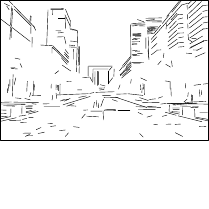}

  \includegraphics[width=0.45\linewidth]{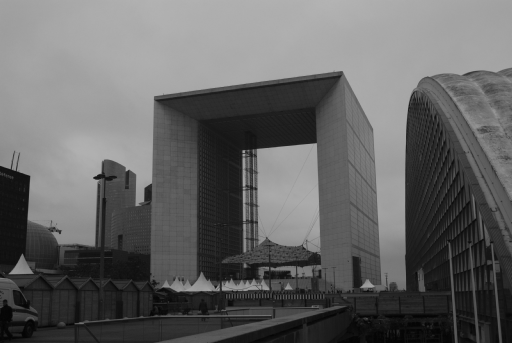}
  \includegraphics[width=0.45\linewidth, trim=0 11.6mm 0 0]{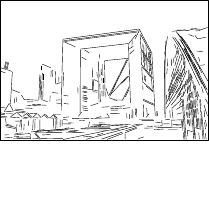}
  \includegraphics[width=0.45\linewidth]{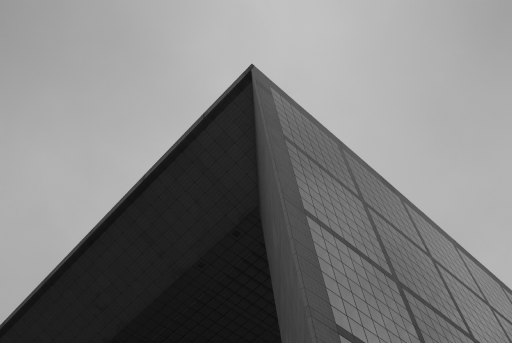}
  \includegraphics[width=0.45\linewidth, trim=0 11.6mm 0 0]{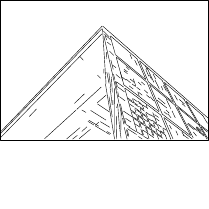}
  \caption[]{Some results of \textit{DSeg}. Note the stability of long segments
    extracted on the square arch building with respect to large scale changes.}
  \label{fig|segmentsResults}
\end{figure}

Figure \ref{fig|segmentsResults} present some results obtained with
\textit{DSeg}.

\subsection{Parameters}
\label{sec|parameters}

The detection process involves the definition of some parameters and
thresholds. Large conservative values for these parameters are easy to specify,
as they do not impact the detection results.  Table \ref{tab|parameters} lists
these parameters and their values -- used in all our trials and for the results
presented in the next sections.

\begin{table}[htb!]
\centerline{
\begin{tabular}{|c|c|c|c|c|c|c|}
  \hline
  $\sigma_a = \sigma_b$ & $\sigma_{x0} = \sigma_{y0}$ & $\delta_t$ & $\tau_{angle}$ & $\tau_{Gmax}$ & $n_o$ & $\sigma_r$ \\
  \hline
  0.05 & 1.0 & 1px & 1.0 - $\sigma_a$ & 10 & 2 & 0.5 \\
  \hline
\end{tabular}}
\caption{Required parameters and associated values}
\label{tab|parameters}
\end{table}

\begin{figure}[tb]
  \centering
  \includegraphics[width=0.4\linewidth]{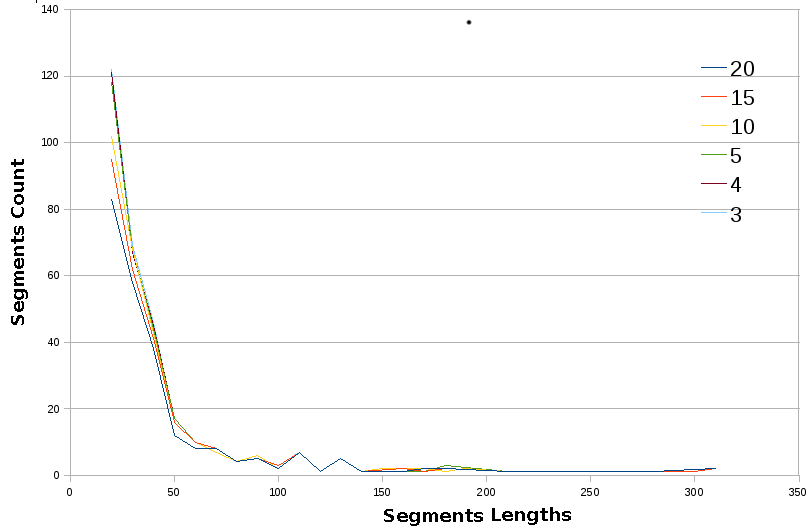}
  \includegraphics[width=0.4\linewidth]{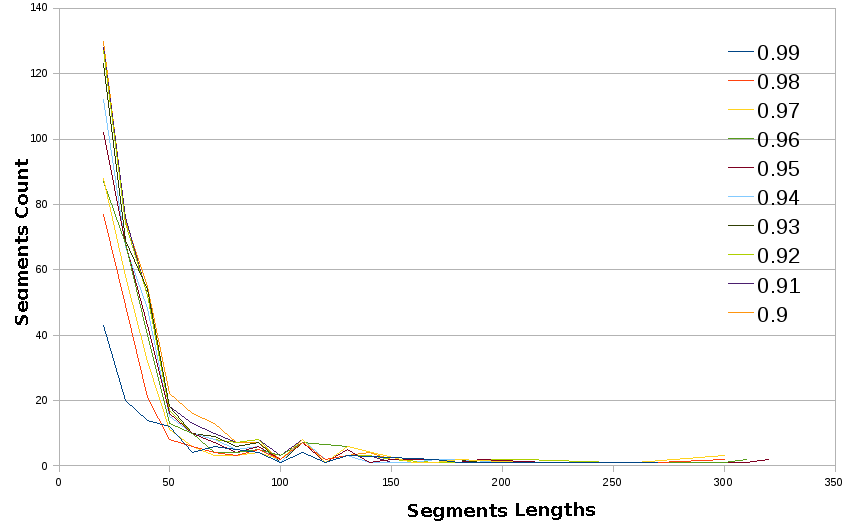}
  \caption[]{Stability of the distribution of the detected segment lengths for
    various values of $\tau_{Gmax}$ (top) and $\tau_{angle}$ (bottom).}
  \label{fig|parametersDetection}
\end{figure}

$\sigma_{x0} = \sigma_{y0}$ indicates how well the position of the local maximum
of a gradient is known, and are set to the image resolution (1 pixel). The two
parameters that are defined empirically are $\sigma_a = \sigma_b$ and
$\tau_{Gmax}$: figure \ref{fig|parametersDetection} shows the lack of influence
of $\tau_{Gmax}$ and $\tau_{angle}$ on the number of detected segments. The
parameter $\tau_{Gmax}$ determines whether low contrast segments will be
extracted or not. For $\tau_{angle}$, a rather large value ($0.95$, $acos(0.95)
= 31\degree)$ is used, so the process is not disturbed by noise and still
ignores drastic changes in orientation. $n_o$ defines the number of gradient
measures in the observation process, and a value of 2 (yielding $2n_o+1=5$
measures) has proven to always suffice.

\section{Hierarchical detection}
\label{sec|hdseg}

\begin{figure}[tb]
  \centering
  \subfigure[Detection on multiple le\-vels.]{
  \includegraphics[width=0.50\linewidth, trim=51mm 120mm 30mm 100mm]{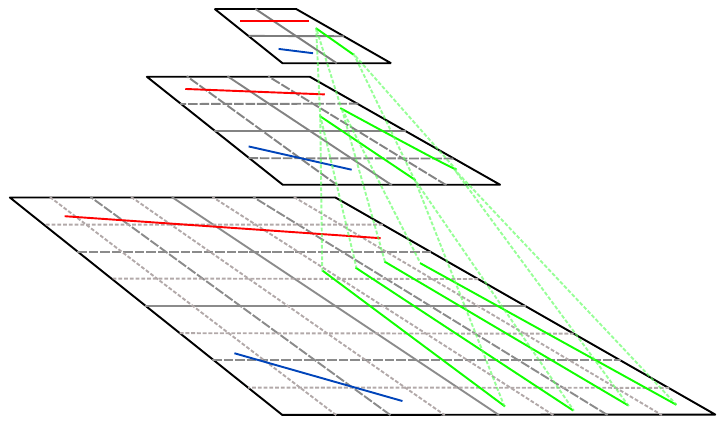}
\label{fig|hd|pyramid} } \subfigure[The red segment is detected at the coarser
level. At the next level, it can correspond to either the green segment, or to
any of the two blue segments.]{
  \includegraphics[width=0.40\linewidth, trim=12mm 240mm 155mm 16mm]{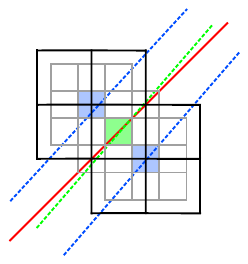}
  \label{fig|hd|nextlevel} }
  \caption[]{Hierarchical detection}
  \label{fig|hd|pyramiddseg}
\end{figure}

This section presents an extension of the detection algorithm, in which segments
are first detected on a scaled down image, and their parameters are then tracked on
the different levels of the pyramid (Figure \ref{fig|hd|pyramiddseg}). Scaling
down the image reduces the noise level, but also yields less precise segments: with
a hierarchical approach, the segment detection is initialized using less noisy
data, but nevertheless attains the precision given by the full size image. Since
scaling down is emulating a distance change (simulating a backward motion of the
camera), another interest is to only detect the segments that are likely to be
visible from a further viewpoint, {\em i.e.} the segments that are more likely
to be stable with respect to scale change.

\begin{figure}[tb]
  \centerline{
    \parbox{0.24\linewidth}{
      \center{\includegraphics[height=0.25\linewidth]{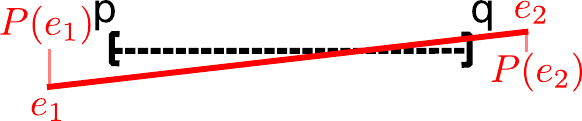}}\\
      \center{ (a)}
    }
    \parbox{0.40\linewidth}{
      \center{\includegraphics[height=0.14\linewidth]{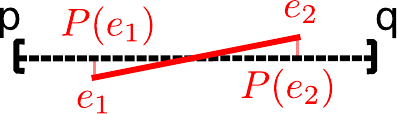}}\\
      \center{ (b) \makeinterval{p}{idx1}, \makeinterval{idx1}{q}}
    }
    \parbox{0.24\linewidth}{
      \center{\includegraphics[height=0.25\linewidth]{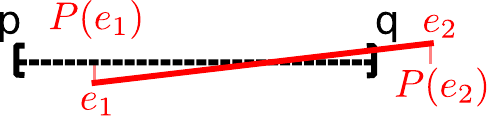}}\\
      \center {(c) \makeinterval{p}{idx1}}
    }
    \parbox{0.24\linewidth}{
      \center{\includegraphics[height=0.25\linewidth]{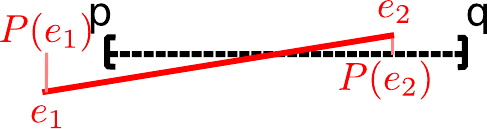}}\\
      \center{\center (d) \makeinterval{idx2}{q}}
    }
  }
  \label{fig|hd|interval3} 
\caption{These figures show how intervals are split after a detection of a
  segment. The indexes $ idx1 < O_j^i P(e_1), \mathbf{u} > $ and $ idx2 = < O_j^i P(e_2), \mathbf{u}> $
  are respectively the distance of the first and second extremity of the newly detected segment to the origin
  (which is defined as one of the extremity of the segment detected at the upper
  level). One situation is not shown here, when $ idx2 < p $ or $ q < idx1 $,
  in which case the interval $ [p,q] $ is not changed.}
  \label{fig|hd|intervals}
\end{figure}

\paragraph*{Detection process.} From the image, a $ n $ levels pyramid is
computed with a scale factor $s_p$. The level $ l^{i = 0} $ correspond to
the bottom of the pyramid (full resolution image), while $ l^{i = n - 1} $
is the level with the lowest resolution. Within this pyramid, the detection
process is handled as follows:

\begin{enumerate}
\item On the top level $ l^{ i = n - 1} $, segments are detected with the
  \textit{DSeg} process (section \ref{sec|dseg}).
\item For each segment $ S_j^i $ detected at level $ l^{i} $, a first estimate
  of the parameters of the same segment $S_j^{i-1}$ at level $l^{i-1}$ is
  obtained by multiplying by $s_p$ the parameters of $s_j^i$. Given $\mathbf{u}$
  the vector director of the segment $S_j^{i-1}$, $ \mathbf{n} $ its normal vector, and $L$ its length, a
  list of intervals is initialized with ${\cal L} = \{[0,L]\}$.
        \begin{enumerate}
        \item Given an interval $I = [p,q] $ from the list $\cal{L}$, look for a maximum of
          gradient in the direction of $\mathbf{n}$, around the point:
                \begin{gather}
                  r = \frac{ p + q}{s_p} \\
                  (x_I, y_I) = (x^{i}, y^{i-1}) + r \mathbf{u}
                \end{gather}           
                $(x^{i-1}, y^{i-1})$ is the origin of a segment $S_j^{i-1}$ at level $i-1$.
              \item If there is a gradient maximum in $ (x_I, y_I) $, $ (x_I,
                y_I) + \mathbf{n} $ or $ (x_I, y_I) - \mathbf{n} $, the segment
                update as described in section \ref{sssec|observation} is applied.
              \item For each segments detected at level $l^{i-1}$, compute the
                perpendicular projection $ P(e_1) $ and $P(e_2)$ of its extremities $e_1$ and
                $e_2$ on $S_j^{i-1}$. Then, using $ idx1 = < O_j^i 
                P(e_1), \mathbf{u} > $ and $ idx2 = < O_j^i P(e_2),
                \mathbf{u}> $, the area where a segment has already been
                detected is removed from the intervals of $\cal{L}$, as explained 
                figure \ref{fig|hd|intervals}.
              \item If no segment is found, the interval $ I = [p,q]$ is split
                into intervals $ [p,r] $ and $ [r, q] $, which are inserted in $\cal{L}$.
          \item Return to step a) until $\cal{L}$ is empty.
        \end{enumerate}
\end{enumerate}


\paragraph*{Parameters.} \textit{Hierarchical DSeg} has only two more parameters
than \textit{DSeg}: the number $ n_p $ of images in the pyramid, and the scale factor $
s_p $ between two images. The scale parameter $ s_p $ should be smaller than 2
($ s_p \leq 2 $), as a larger values would imply the search of a gradient
maximum around more points in step 2(b). In the following, we use $ s_p = 2 $
and $ n_p = 3 $.

\begin{figure}[tb]
  \centering
  \includegraphics[width=0.45\linewidth]{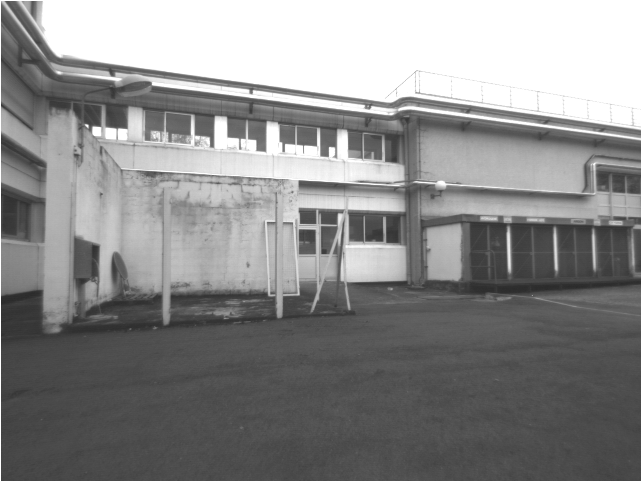}
  \includegraphics[width=0.45\linewidth, trim=0 8.8mm 0 0]{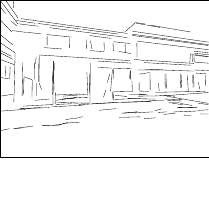}
  \includegraphics[width=0.45\linewidth]{Distance1.png}
  \includegraphics[width=0.45\linewidth, trim=0 11.6mm 0 0]{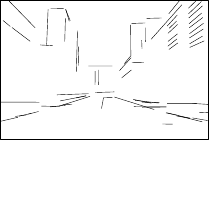}

  \includegraphics[width=0.45\linewidth]{Distance2.png}
  \includegraphics[width=0.45\linewidth, trim=0 11.6mm 0 0]{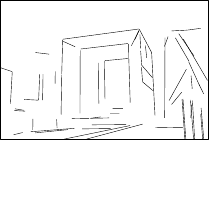}
  \includegraphics[width=0.45\linewidth]{Distance3.png}
  \includegraphics[width=0.45\linewidth, trim=0 11.6mm 0 0]{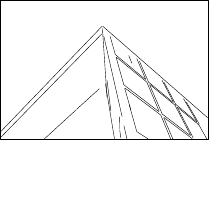}
  \caption[]{Line segments extracted with \textit{Hierarchical DSeg}. Less
    segments are detected than with the \textit{DSeg} (compare with figure
    \ref{fig|segmentsResults}).}
  \label{fig|hdsegsegmentsResults}
\end{figure}

\begin{figure*}[htb!]
  \centering
  \includegraphics[width=0.32\linewidth]{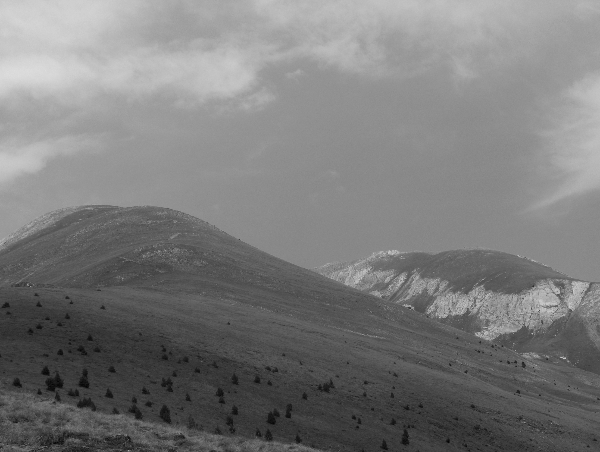}
  \includegraphics[width=0.32\linewidth]{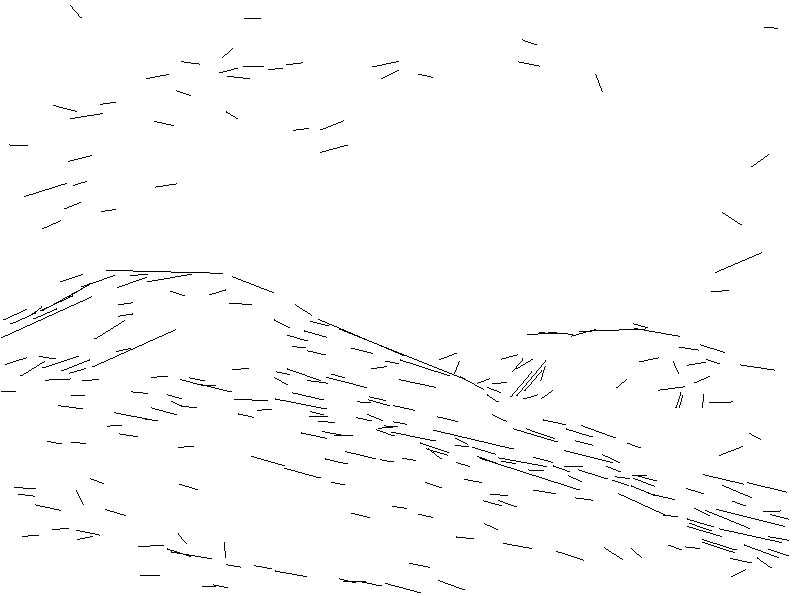}
  \vspace{2mm}
  \includegraphics[width=0.32\linewidth]{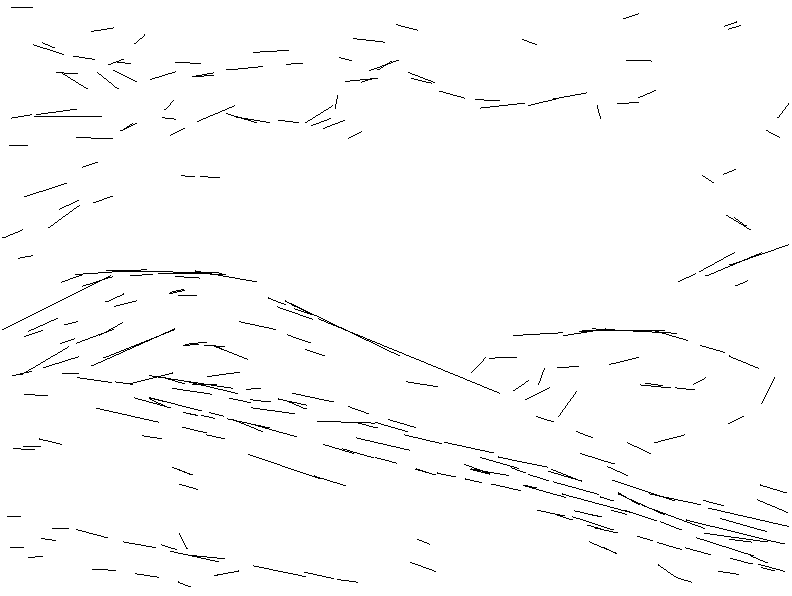}
  \includegraphics[width=0.32\linewidth]{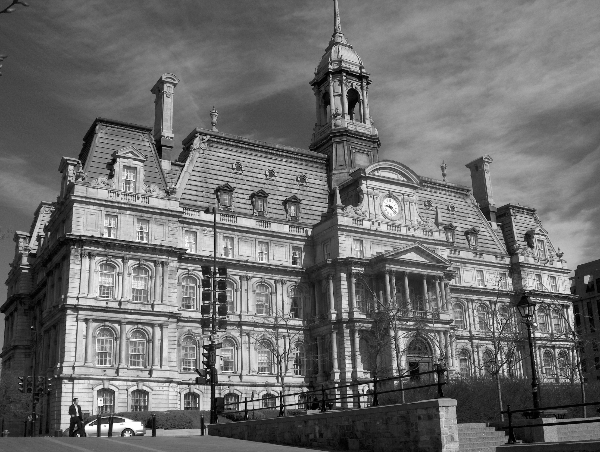}
  \includegraphics[width=0.32\linewidth]{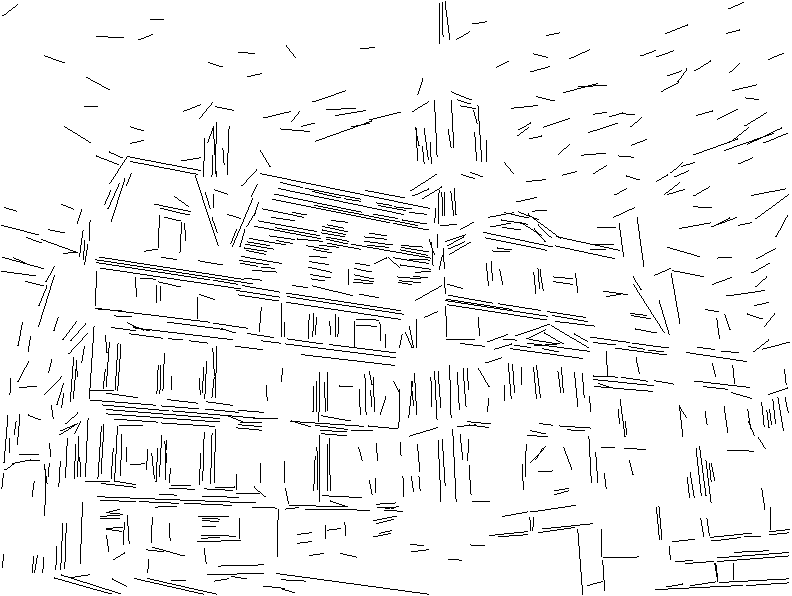}
  \vspace{2mm}
  \includegraphics[width=0.32\linewidth]{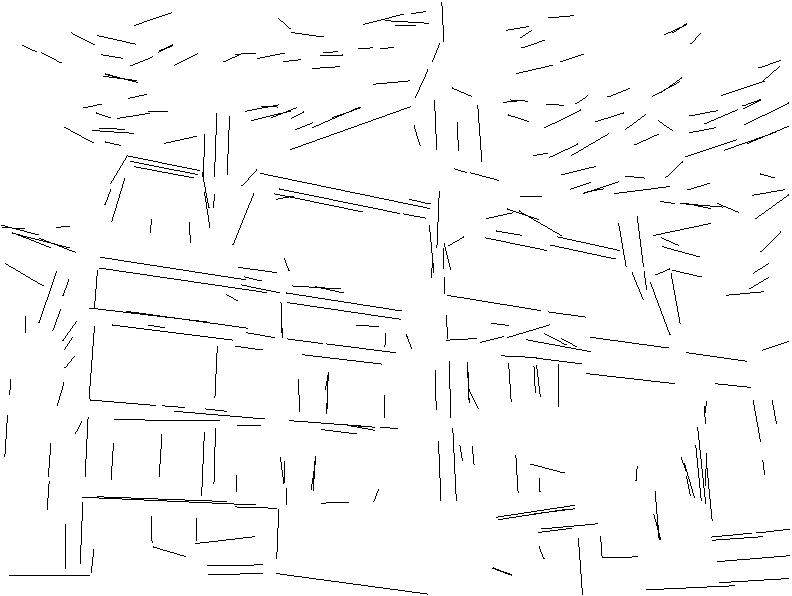}
  \includegraphics[width=0.32\linewidth]{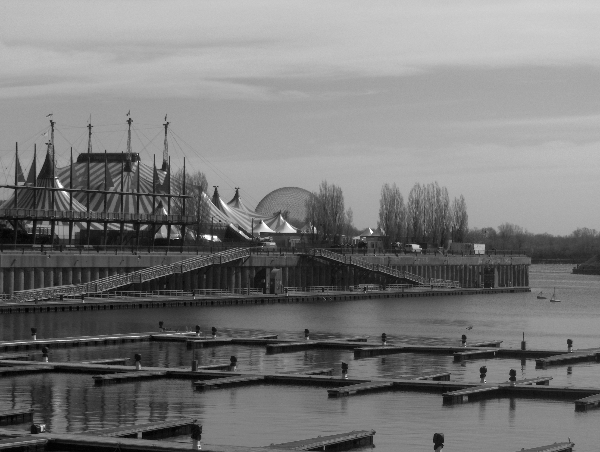}
  \includegraphics[width=0.32\linewidth]{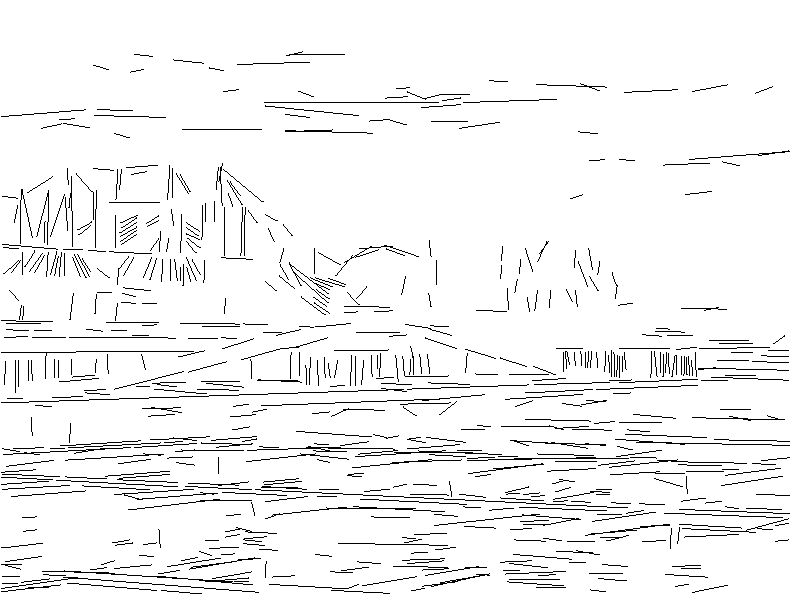}
  \vspace{2mm}
  \includegraphics[width=0.32\linewidth]{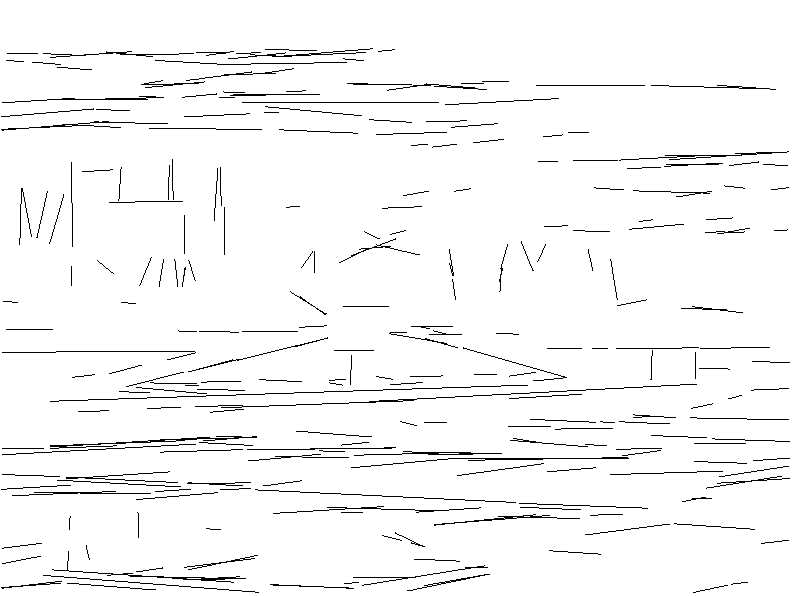}
  \includegraphics[width=0.32\linewidth]{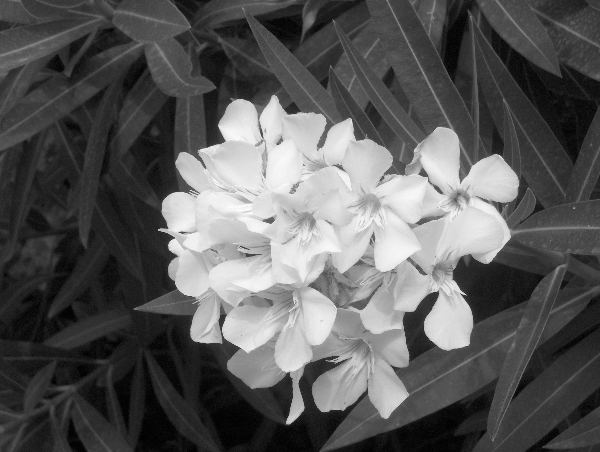}
  \includegraphics[width=0.32\linewidth]{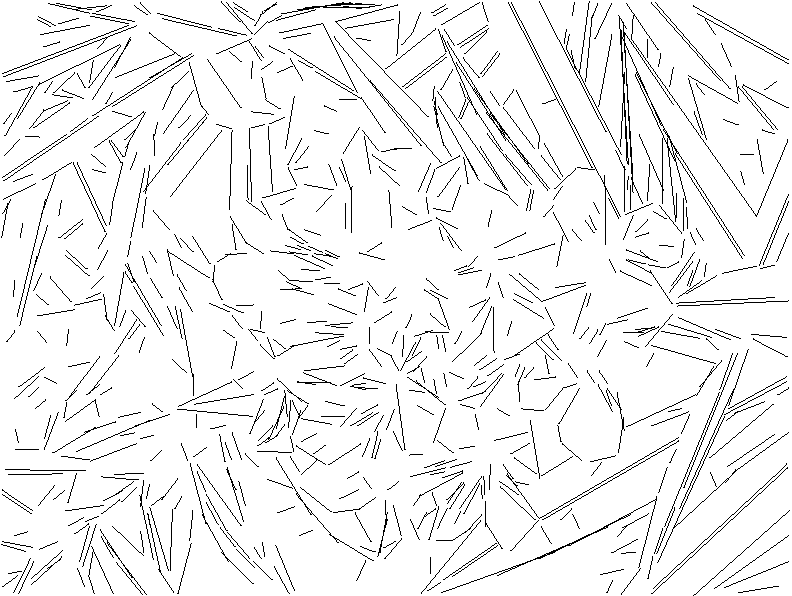}
  \vspace{2mm}
  \includegraphics[width=0.32\linewidth]{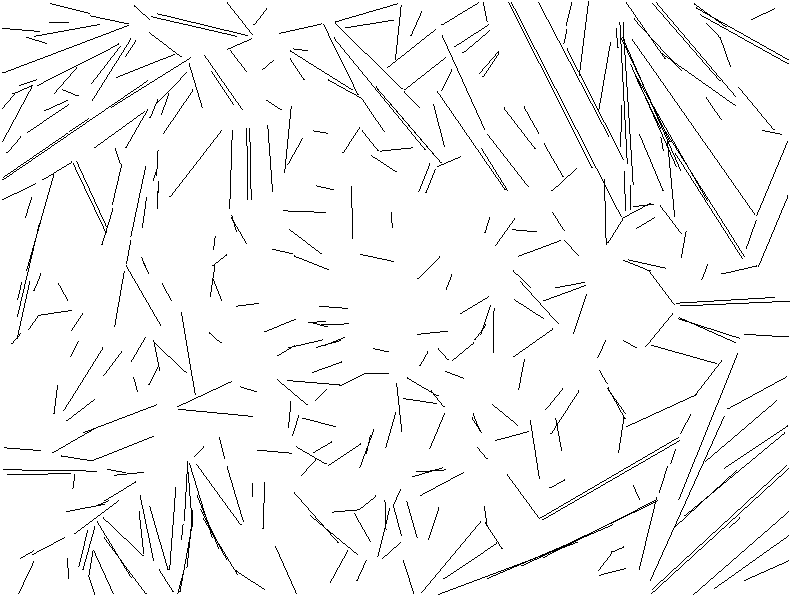}
  \includegraphics[width=0.32\linewidth]{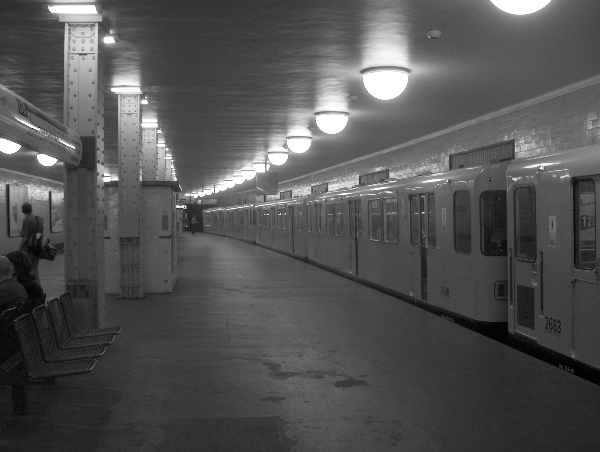}
  \includegraphics[width=0.32\linewidth]{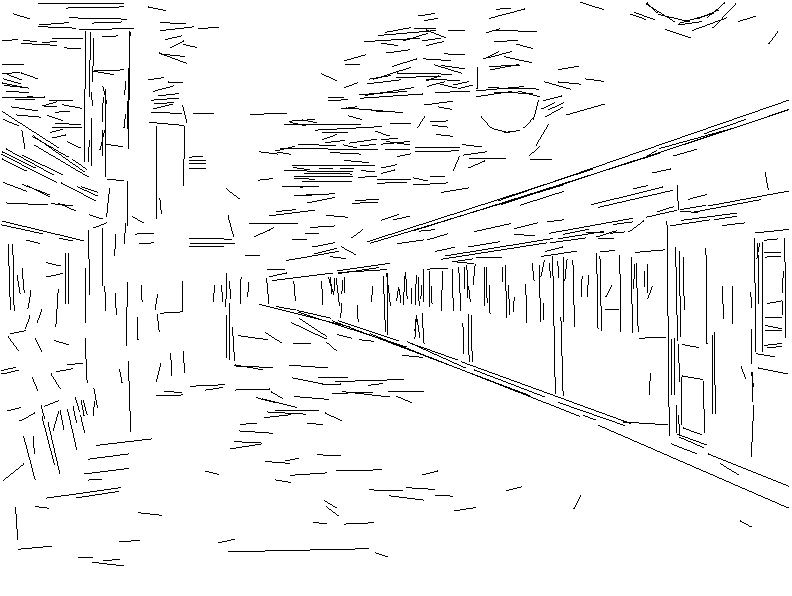}
  \vspace{2mm}
  \includegraphics[width=0.32\linewidth]{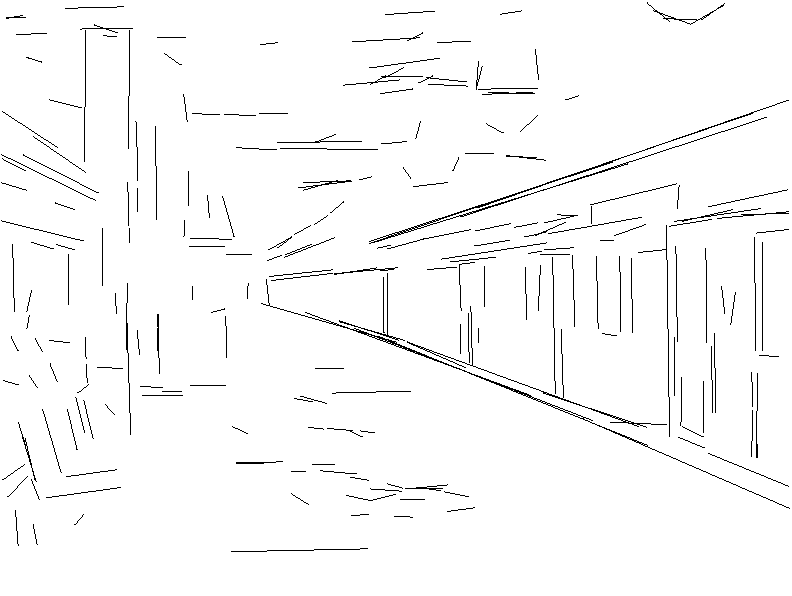}
  \caption[]{Segments extracted with \textit{DSeg} (center) and \textit{Hierarchical DSeg} (right).}
  \label{fig|segmentsResultsMore}
\end{figure*}

\begin{figure*}[htb!]
  \includegraphics[width=0.32\linewidth]{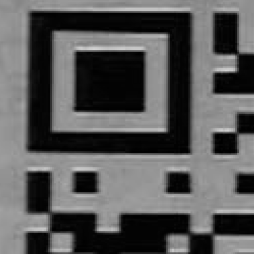}
  \includegraphics[width=0.32\linewidth]{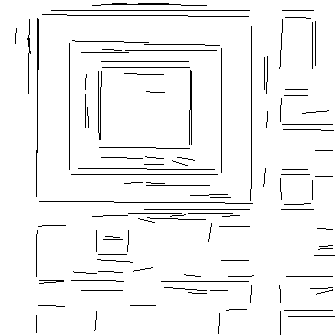}
  \includegraphics[width=0.32\linewidth]{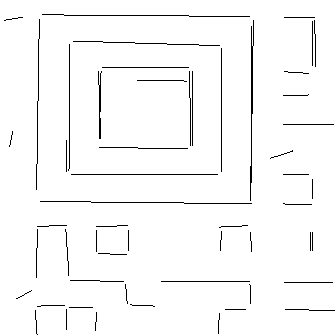}
  \vspace*{2mm}
  \includegraphics[width=0.32\linewidth]{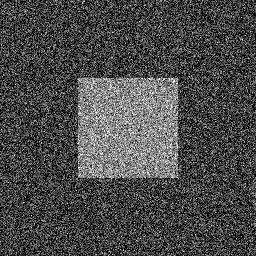}
  \includegraphics[width=0.32\linewidth]{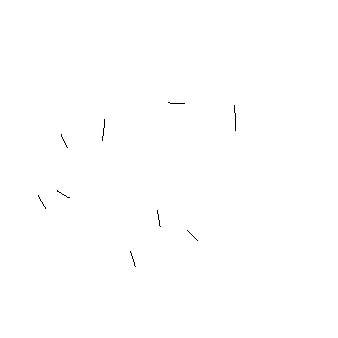}
  \includegraphics[width=0.32\linewidth]{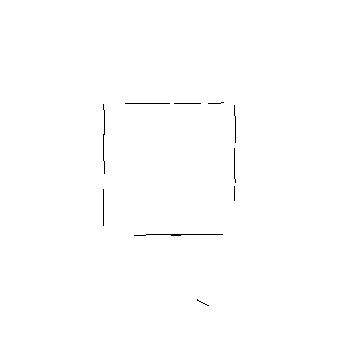}
  \vspace*{2mm}
  \includegraphics[width=0.32\linewidth]{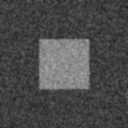}
  \includegraphics[width=0.32\linewidth]{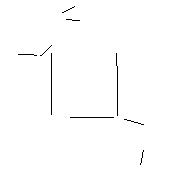}
  \includegraphics[width=0.32\linewidth]{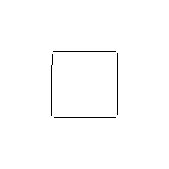}
  \caption[]{Segments extracted with \textit{DSeg} (center) and \textit{Hierarchical DSeg} (right).}
  \label{fig|segmentsResultsEvenEvenMore}
\end{figure*}

\paragraph*{Results.} Figure \ref{fig|hdsegsegmentsResults} shows the result of a
detection on different images, and figures \ref{fig|segmentsResultsMore} and
\ref{fig|segmentsResultsEvenEvenMore} compare the results of \textit{DSeg} and
\textit{Hierarchical DSeg} on various images.

The results of figure~\ref{fig|segmentsResultsMore} show that both algorithms
successfully extract line segments on images taken under various conditions and
in different types of environments. Rather obviously, urban environment gives
more interesting results that natural
landscapes. Figure~\ref{fig|segmentsResultsEvenEvenMore} shows the extraction on
synthetic images, and it demonstrates the stronger robustness of the hierarchical
detection with respect to noise. The figures are here only to demonstrate the
effect of noise on highly contrasted images, where segment location is well defined,
other algorithms such as \cite{NGUYEN-PR-2011} are better suited for extracting contours
on that type of images.

\section{Comparative analysis}
\label{sec|comparision}

Here we present a comparison of our approaches (denoted \textit{DSeg} and
\textit{Hierarchical DSeg}), with the Probabilistic Hough transformation
approach (\cite{GUO-2008}, denoted \textit{Hough}, OpenCV implementation), the
segments chaining approach (\cite{ETEMADI-1992}, denoted \textit{Chaining}), our
implementation), and the approach presented in \cite{GROMPONE-2008} (denoted
\textit{LSD}, using Grompone's implementation).

\subsection{Number and length of detected segments}

\begin{figure}[htb!]
  \centering
  \subfigure[Image 640x480]{
  \includegraphics[width=0.3\linewidth]{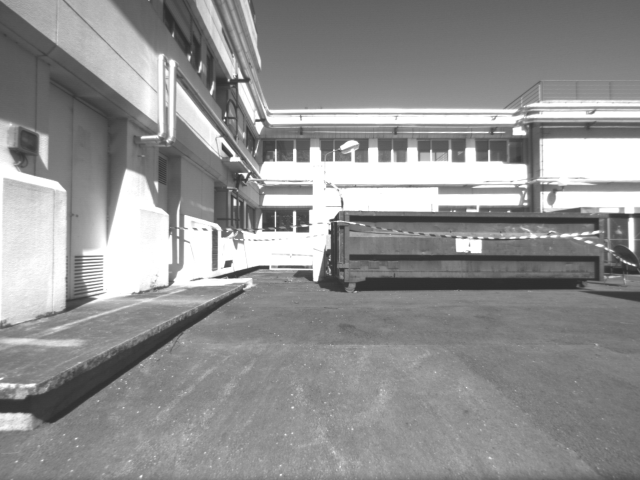} }
  \subfigure[Canny]{
  \includegraphics[width=0.3\linewidth]{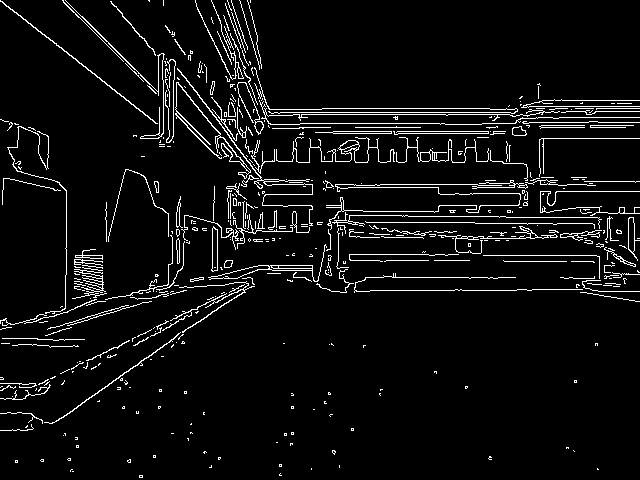} }
  \subfigure[Probabilistic Hough (82ms)]{
  \includegraphics[width=0.3\linewidth, trim=0 9mm 0 0]{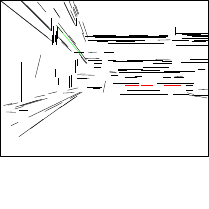} }
  \subfigure[Chaining (60ms)]{
  \includegraphics[width=0.3\linewidth, trim=0 9mm 0 0]{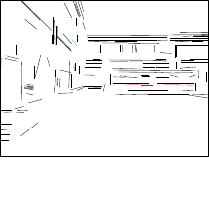} }
  \subfigure[LSD (86ms)]{
  \includegraphics[width=0.3\linewidth, trim=0 9mm 0 0]{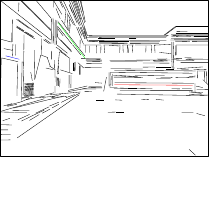} }
  \subfigure[DSeg (168ms)]{
  \includegraphics[width=0.3\linewidth, trim=0 9mm 0 0]{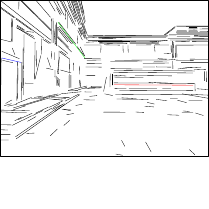} }
  \subfigure[Hierarchical DSeg (181ms)]{
  \includegraphics[width=0.3\linewidth, trim=0 9mm 0 0]{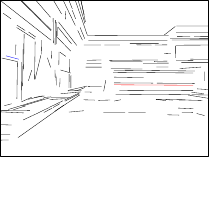} }
  \subfigure[Segment length]{
  \label{fig|segmentsComparisonHistoHC}
  \includegraphics[width=0.3\linewidth]{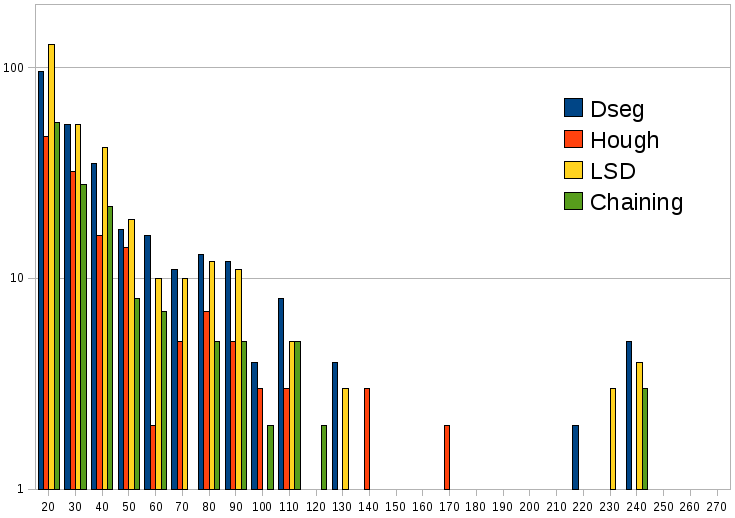} }
  \caption[]{Line segments extracted on a well contrasted image (parameters for Chaining
             and Hough were manually optimised)}
  \label{fig|segmentsComparisonHC}
\end{figure}

\begin{figure}[htb!]
  \centering
  \subfigure[Image 512x384]{
  \includegraphics[width=0.3\linewidth]{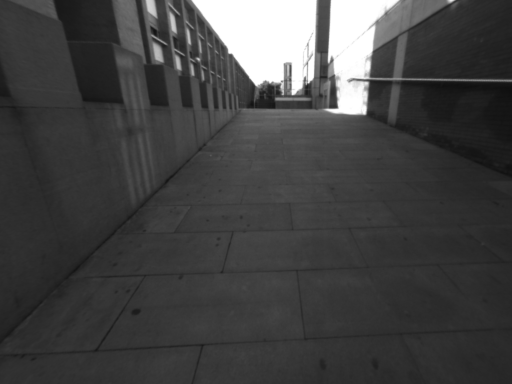} }
  \subfigure[Canny]{
  \includegraphics[width=0.3\linewidth]{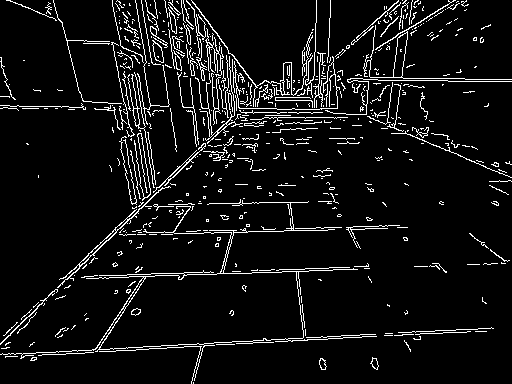} }
  \subfigure[Probabilistic Hough (70ms)]{
  \includegraphics[width=0.3\linewidth, trim=0 9mm 0 0]{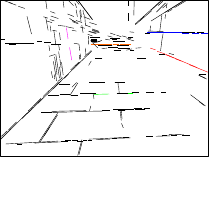} }
  \subfigure[Chaining (30ms)]{
  \includegraphics[width=0.3\linewidth, trim=0 9mm 0 0]{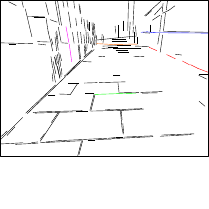} }
  \subfigure[LSD (151ms)]{
  \includegraphics[width=0.3\linewidth, trim=0 9mm 0 0]{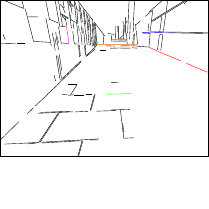} }
  \subfigure[DSeg (78ms)]{
  \includegraphics[width=0.3\linewidth, trim=0 9mm 0 0]{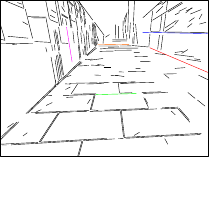} }
  \subfigure[Hierarchical DSeg (83ms)]{
  \includegraphics[width=0.3\linewidth, trim=0 9mm 0 0]{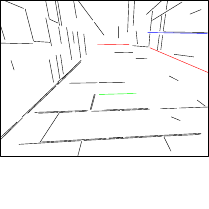} }
  \subfigure[Segment length]{
  \label{fig|segmentsComparisonHistoLC}
  \includegraphics[width=0.3\linewidth]{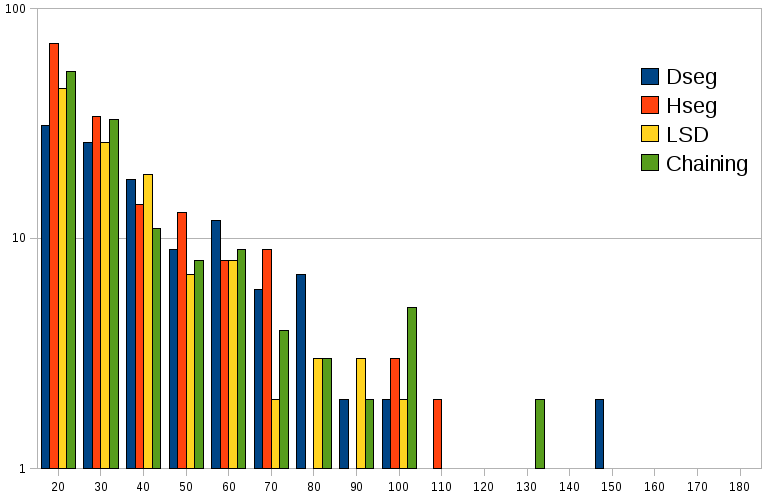} }
  \caption[]{Line segments extracted on a low contrast image (parameters for Chaining
             and Hough were manually optimised).}
  \label{fig|segmentsComparisonLC}
\end{figure}

First note that with the \textit{Hough} and \textit{Chaining} approaches, it has
been necessary to adjust the parameters for each image to obtain good results --
especially the two thresholds on the results of the Canny filter, which need to
be tuned depending on the contrast and noise level of the image. Whereas for the
\textit{LSD} approach and ours, it has not been necessary to change any
parameters.

Figure \ref{fig|segmentsComparisonLC} shows the line segments detected by the
four approaches, with the associated computed times\footnote{Assessed on an 2.2
  GHz Intel processor} -- only segments longer than 20 pixels are
shown. \textit{DSeg} extract more segments than the three other approaches, and
especially longer segments -- as highlighted by red and blue segments in the
figures. Note that on more contrasted images (Figure
\ref{fig|segmentsComparisonHC}), \textit{DSeg} and \textit{LSD} behave quite
similarly, and better than the \textit{Hough} and \textit{Chaining} approaches.

The computation times show that all approach take a rather similar amount of
time, with a slight advantage for \textit{Hough} and \textit{Chaining}. Note
that \textit{Hierarchical DSeg} does not provide any performance enhancement
compared to \textit{DSeg}: the steps of both algorithms that consume most of the
time is the detection and incorporation of support points, and in
\textit{Hierarchical DSeg}, these steps are applied several times at different
scales.

More quantitative results on the number and length of the detected segments are
provided by the histograms of figures \ref{fig|segmentsComparisonHistoHC} and
\ref{fig|segmentsComparisonHistoLC}: \textit{DSeg} finds the longest segments,
and in total a larger number of segments. Logically, \textit{Hierarchical DSeg}
finds less segments.

\subsection{Sensitivity analysis}
\label{ssec|sensitivity}

To assess the robustness of the algorithm with respect to image noise and 
illumination changes, an analysis of the {\em repeatability} of the detected
segments is presented by adding noise to an image, and on a sequence of images
taken during 24 hours with a still camera.


\subsubsection{Assessing the segment repeatability}

\begin{figure}[htb!]
\centering
\includegraphics[width=0.6\linewidth, trim=0 160mm 150mm 0mm]{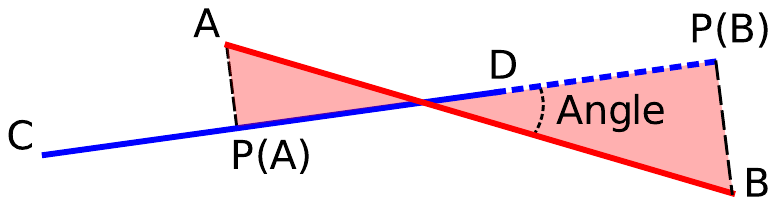}
\caption[]{Distance between two lines segments. Here the overlap is $ \mathcal{R}(P_{CD}(AB),CD) =
\frac{|P_{CD}(A)D|}{|P_{CD}(B)C|}$.
\label{figure|distancesegment}
}
\end{figure}

To be able to automatically compute the repeatability of the detection of
segments with a fixed camera acquiring images of a static scene, it is necessary
to be able to compute which segment of the reference frame corresponds to the
segment detected at a given time. To do this, a similarity measure is defined,
that uses the area contained between the two segments, weighted by their length,
angle and overlap (figure \ref{figure|distancesegment}):

%
%
%
%
%

\begin{equation}
sim(AB, CD) = \dfrac{\mathcal{A}(AB, P_{CD}(AB))}{|P_{CD}(AB)| \cdot \mathcal{R}(P_{CD}(AB),CD) \cdot |cos\widehat{(AB,CD)}|} \\
\end{equation}

where: $P_{CD}(AB)$ is the projection of the segment $AB$ on $CD$,
$\mathcal{A}(AB, P_{CD}(AB))$ is the area between the segment $AB$ and its
projection on the segment $CD$, $|P_{CD}(AB)|$ is the length of the projected
segment, $\mathcal{R}(P_{CD}(AB),CD)$ measures the overlap between the
projection and the segment $CD$, and $\widehat{(AB,CD)}$ is the angle between
the two segments. A small value of the similarity would indicate that the two
segments $AB$ and $CD$ are very similar to each other. Note that this similarity
measure is not scale invariant, but this is not an issue, since it is only used
to assess the repeatability of the detection process.

The similarity measure is used to define the distance:

\begin{equation}
d(AB, CD) = sim(AB, CD) + sim(CD, AB) \label{eq|distance}
\end{equation}

which allows finding the segment of the reference image closest to
the segment in the current image. This distance is not a distance in the
mathematical sense, since $ d(AB, CD) = 0 $ means that $ \mathcal{A}(AB,
P(AB,CD)) = 0 $, which can happen when $A$, $B$, $C$ and $D$ are aligned, or
$AB$ and $CD$ are orthogonal (in which case $cos(angle(AB,CD)) = 0$). It is
however commutative, and if $d(AB, CD) < d(AB, EF)$, then $CD$ is closer to $AB$
than to $EF$ (and the fact that is it not strictly speaking a distance is not
an issue since it is used here only for comparison purposes).

Given two sets of segments $ \Sigma_{ref} $ and $ \Sigma_t $, the
repeatability is defined as the number of segments $ S^i_{ref} \in \Sigma_{ref} $ that
satisfy the following conditions:
\begin{gather}
  \exists S^j_t \in \Sigma_t / \forall k \neq j, d(S^i_{ref}, S^j_t) < d(
  S^i_{ref},
  S^k_t) \\
  \forall l \neq i, d(S^i_{ref}, S^j_t) < d( S^l_{ref}, S^j_t) \\
  d(S^i_{ref}, S^j_t) < \tau_{dist}
\end{gather}

Since it is always possible to find a closest segment, a threshold $ \tau_{dist}
$ is used to eliminate segments that are not close to the current segment. The
threshold has been empirically set as $ \tau_{dist} = 50 $.

\subsubsection{Sensitivity to noise}

To test the sensitivity of the algorithms to noise, additive and
multiplicative noises are applied to the luminance value of a reference image $
I_{ref} $:
\begin{equation}
I(i) = R(\sigma_m^i) I_{ref} + R(\sigma_a^i)
\label{equ|noise}
\end{equation}

where $ R(\sigma) $ is a random number generator with a Gaussian distribution of
standard deviation $ \sigma $ and a null mean. The following parameters were
used:

\begin{equation}
  \sigma_n^i = \sigma_a^i = 5i \label{equ|sigmanoise}
\end{equation}

where $i$ is the number of the test frame ($i = 0$ is the original source
image).

\begin{figure}[htb!]
\subfigure[Reference image]{
  \includegraphics[width=0.28\linewidth, trim=5mm 9mm 60mm 5mm]
                  {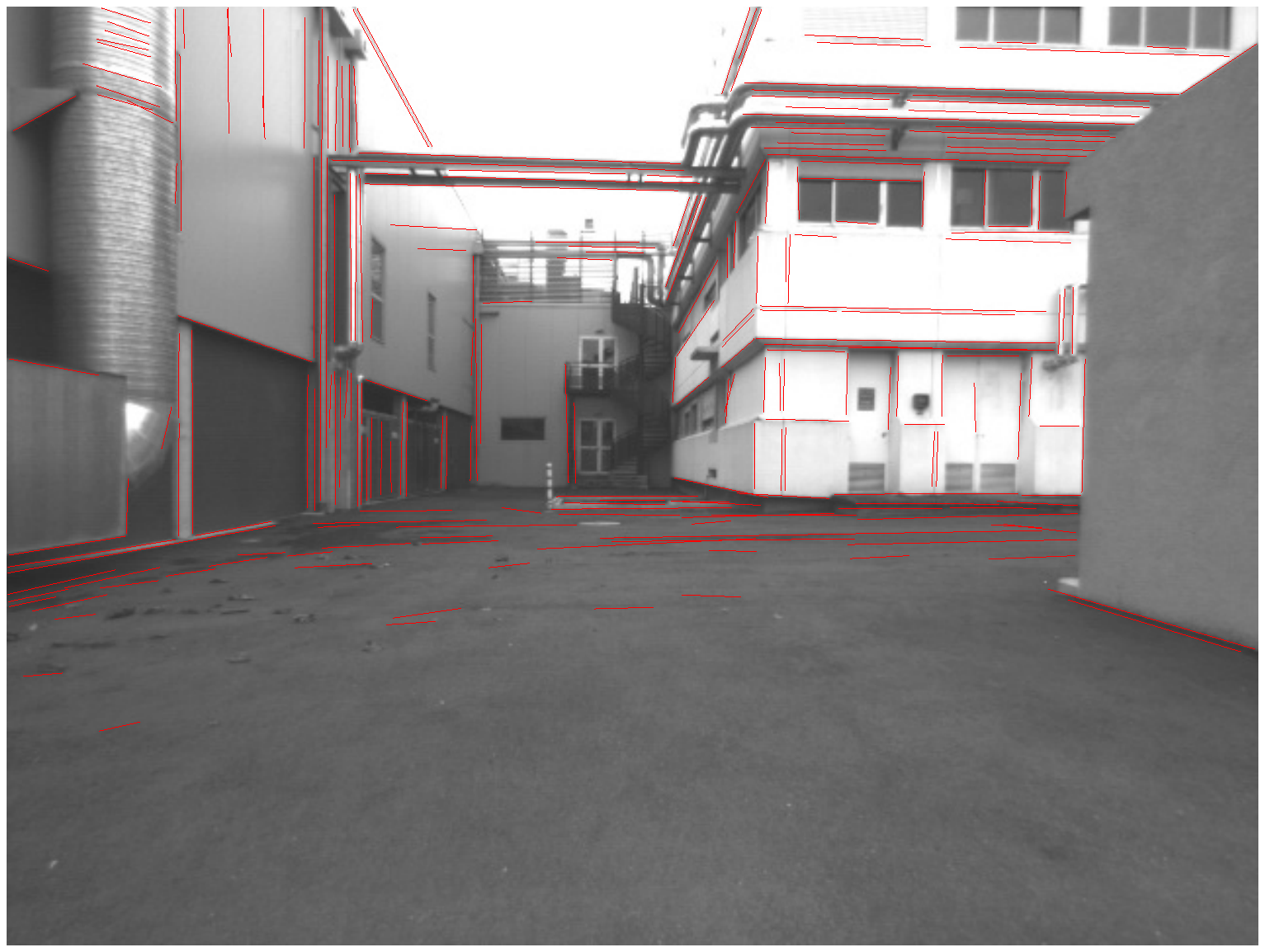} }
\hfill
\subfigure[Image half noise]{
  \includegraphics[width=0.28\linewidth, trim=5mm 9mm 60mm 5mm]
                  {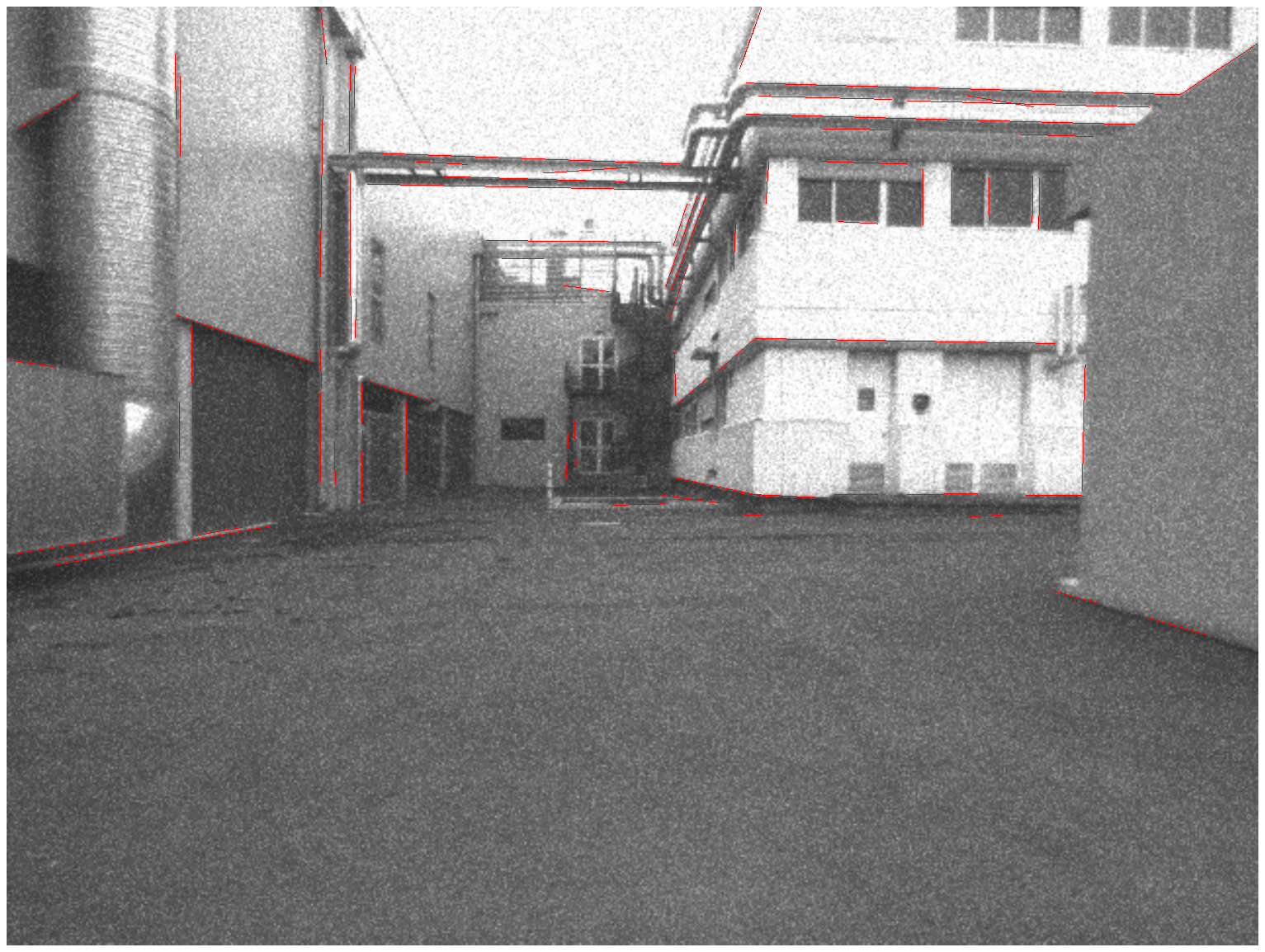} }
\hfill
\subfigure[Image full noise]{
  \includegraphics[width=0.28\linewidth, trim=5mm 9mm 60mm 5mm]
                  {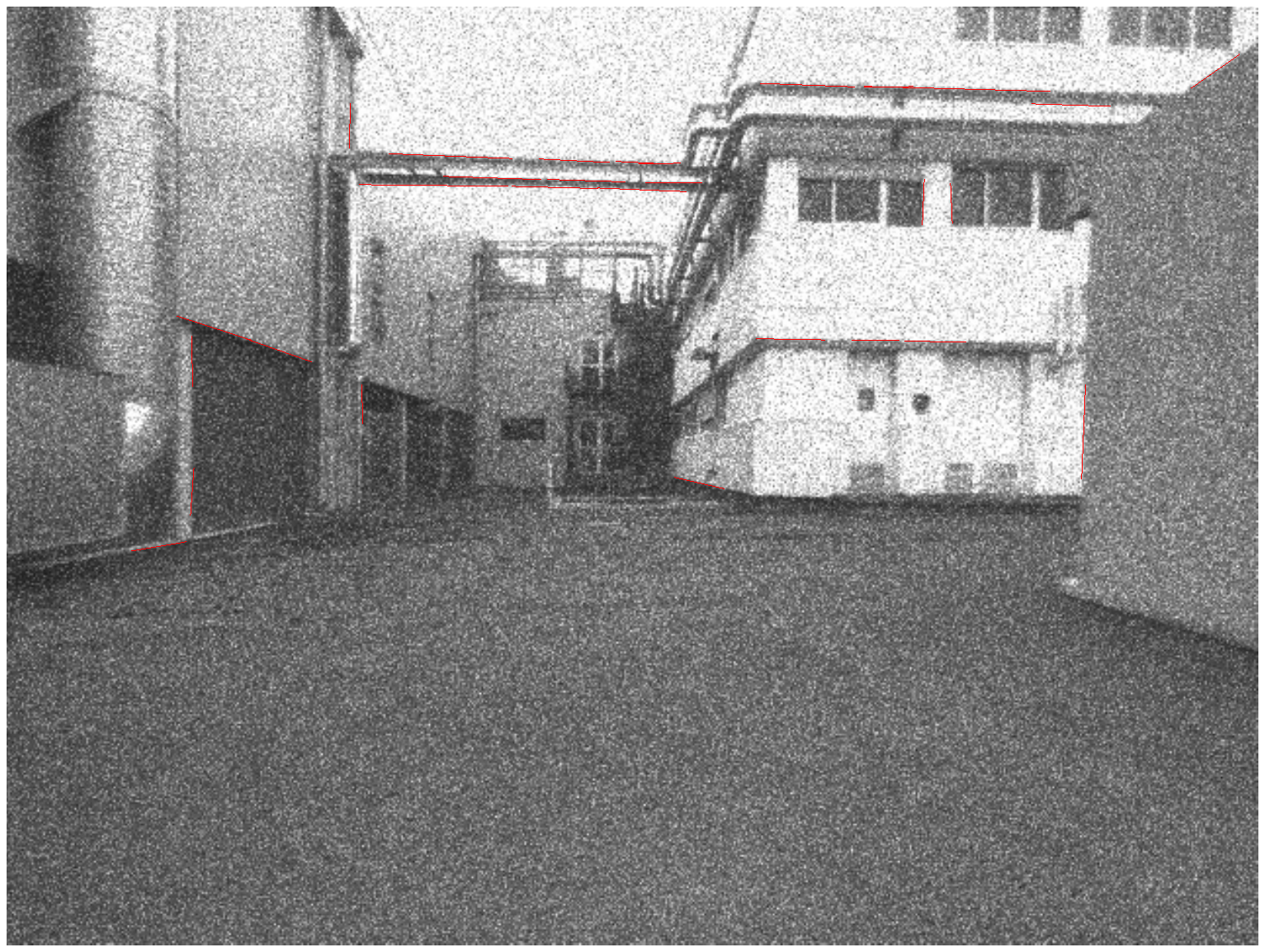} }

\includegraphics[width=\linewidth, trim=0 195mm 50mm 0]{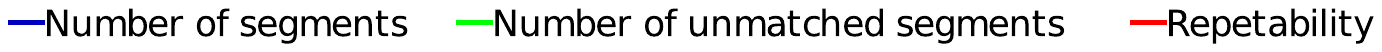}

\subfigure[$ \tau_{Gmax} = 0 $]{
  \includegraphics[width=0.49\linewidth]
                  {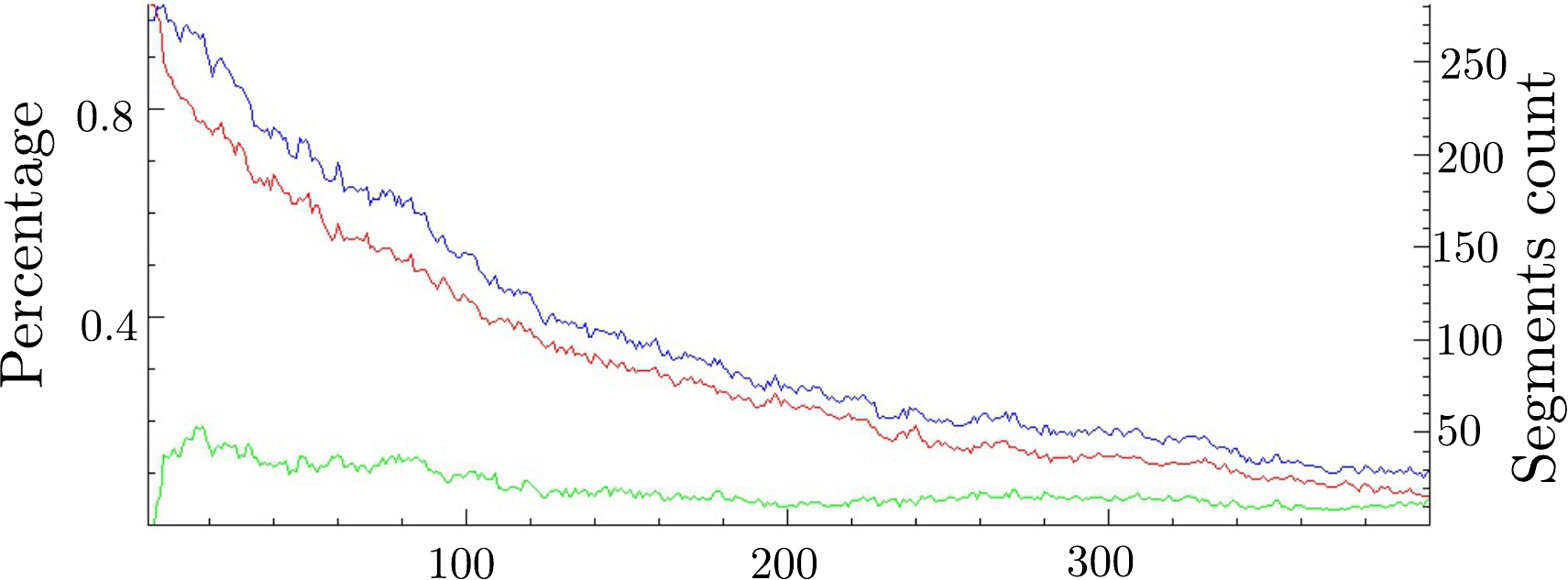} }
\hfill
\subfigure[$ \tau_{Gmax} = 10 $]{
  \includegraphics[width=0.49\linewidth]
                  {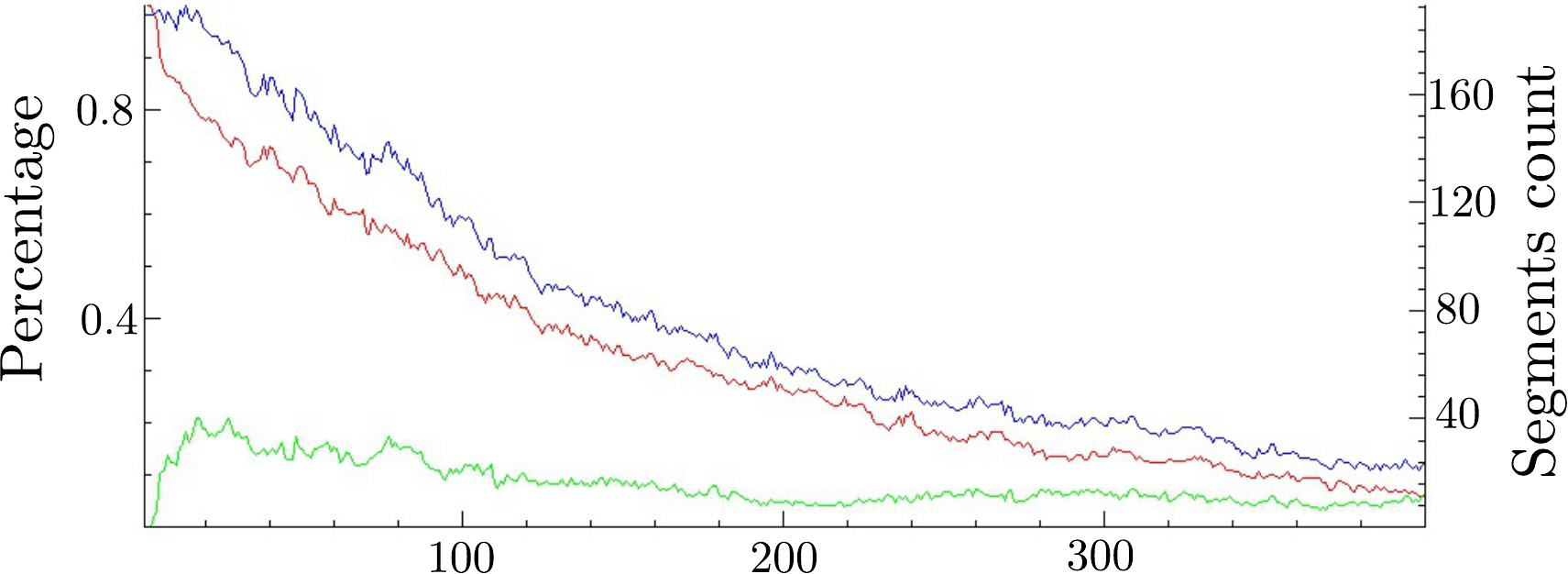} }
\subfigure[$ \tau_{Gmax} = 20 $]{
  \includegraphics[width=0.49\linewidth]
                  {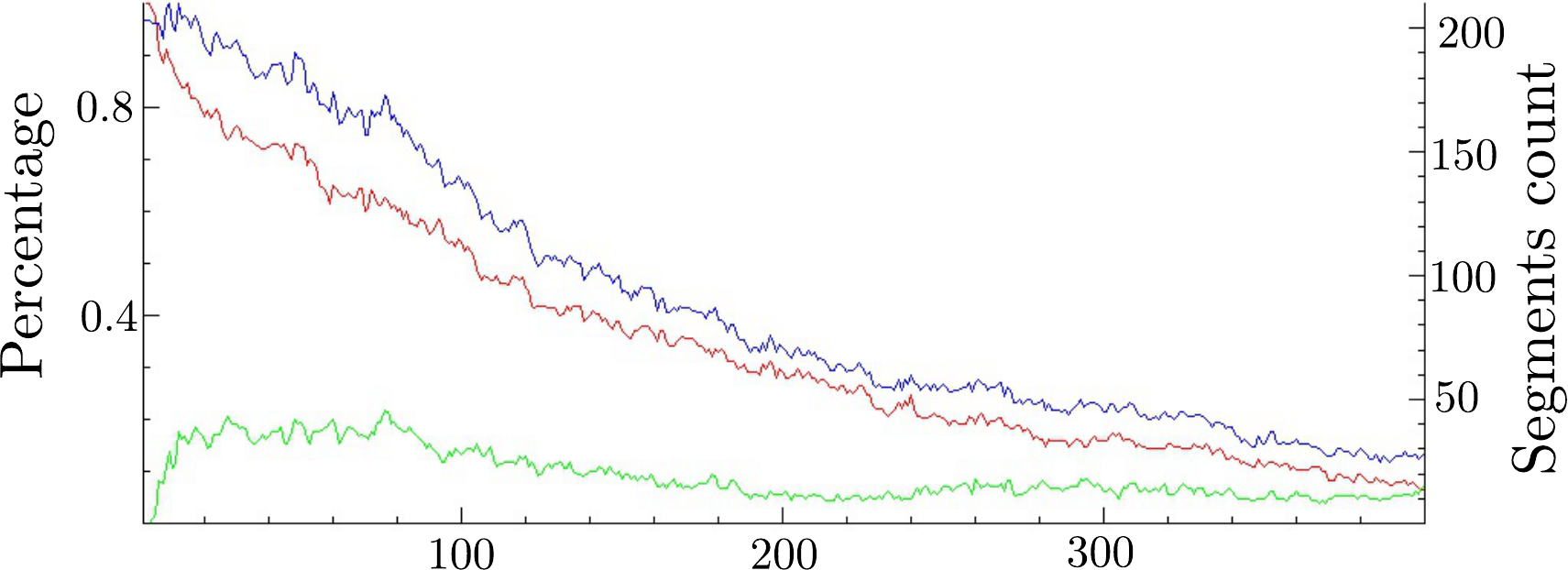} }
\hfill
\subfigure[$ \tau_{Gmax} = 30 $]{
  \includegraphics[width=0.49\linewidth]
                  {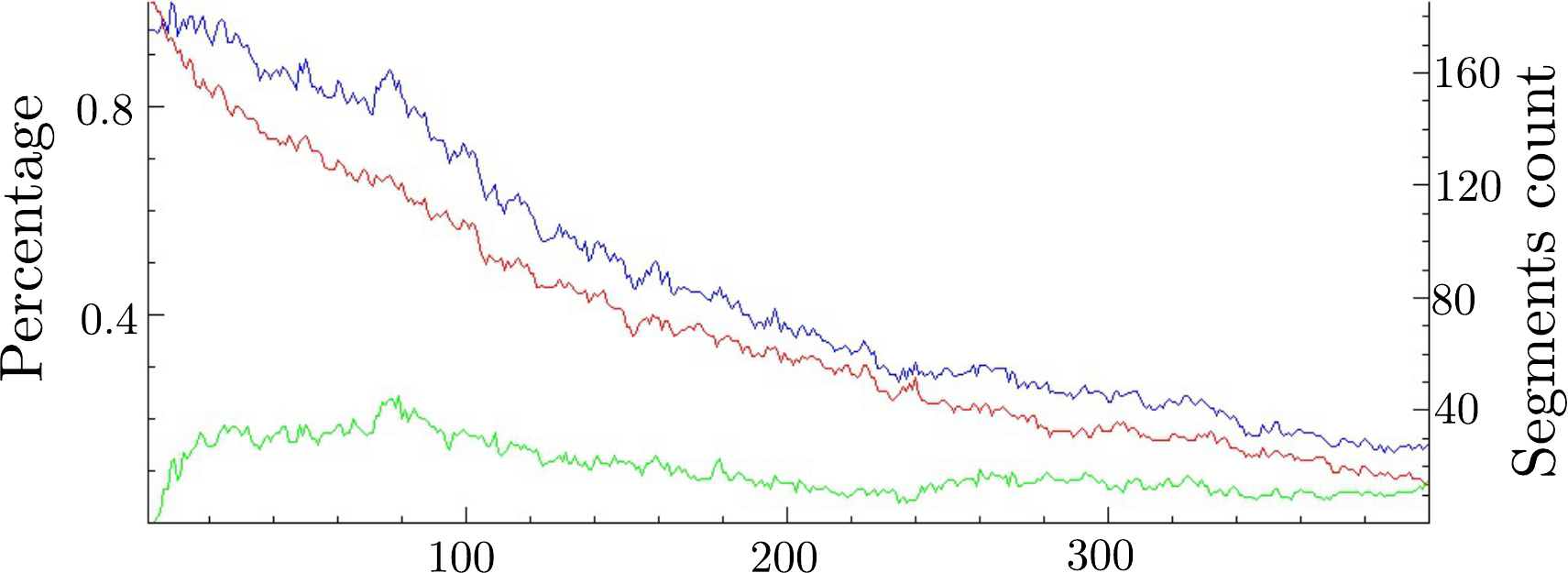} }
\caption{Sensitivity to noise for DSeg: evolution of the repeatability of the
  detection as a function of the image noise. The horizontal axis represents
  the amount of noise added to the image following equation~\ref{equ|sigmanoise}.}
  
\DONE{11) The caption of Figs.14-18 should explain better the figures. For example,
the x-axis are not explained.
12) In Fig.14 the line segment count axis range is very different for the
different graphs, giving the idea that the results a very similar for different
values of tau\_Gmax, while it is only the percentage that is similar.}
  
\label{fig|segmentsAddedNoise}
\end{figure}

\paragraph*{DSeg.} Figure \ref{fig|segmentsAddedNoise} shows the number of
segments detected by \textit{DSeg} for different values of $ \tau_{Gmax} $ (the
threshold used to select segment seeds), as a function of the image noise. It
shows that $ \tau_{Gmax} $ has no influence on the repeatability of segment
extraction, and that the algorithm is robust with respect to a significant level
of noise.

\DONE{Les légendes des courbes de la figure \ref{fig|segmentsAddedNoise} sont
  illisibles !}

\begin{figure}[htb!]
\centering \includegraphics[width=0.60\linewidth]{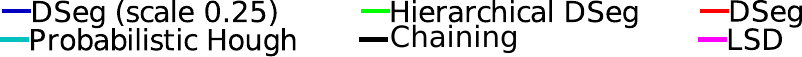}
\subfigure[Number of detected segments]{
\centering 
  \includegraphics[width=0.68\linewidth]
      {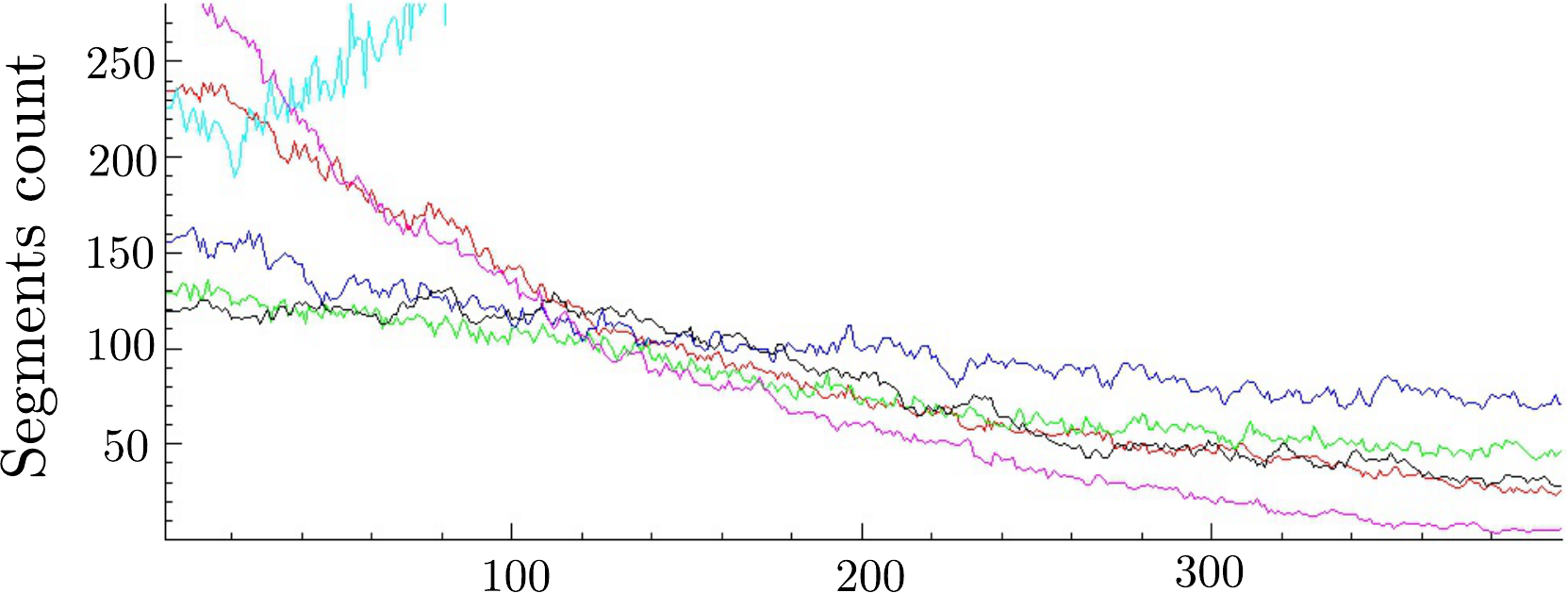} }
\subfigure[Repeatability]{
\centering
  \includegraphics[width=0.68\linewidth]
      {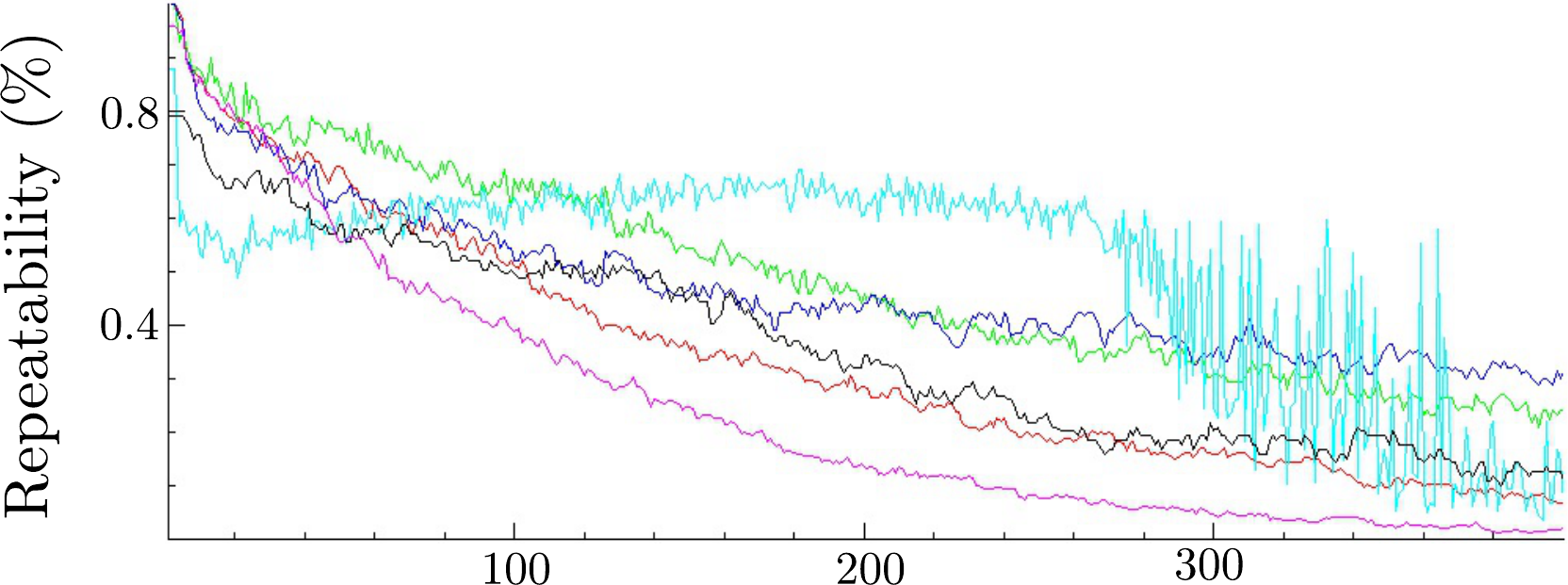} }
\subfigure[Number of unmatched segments]{
\centering 
  \includegraphics[width=0.68\linewidth]
      {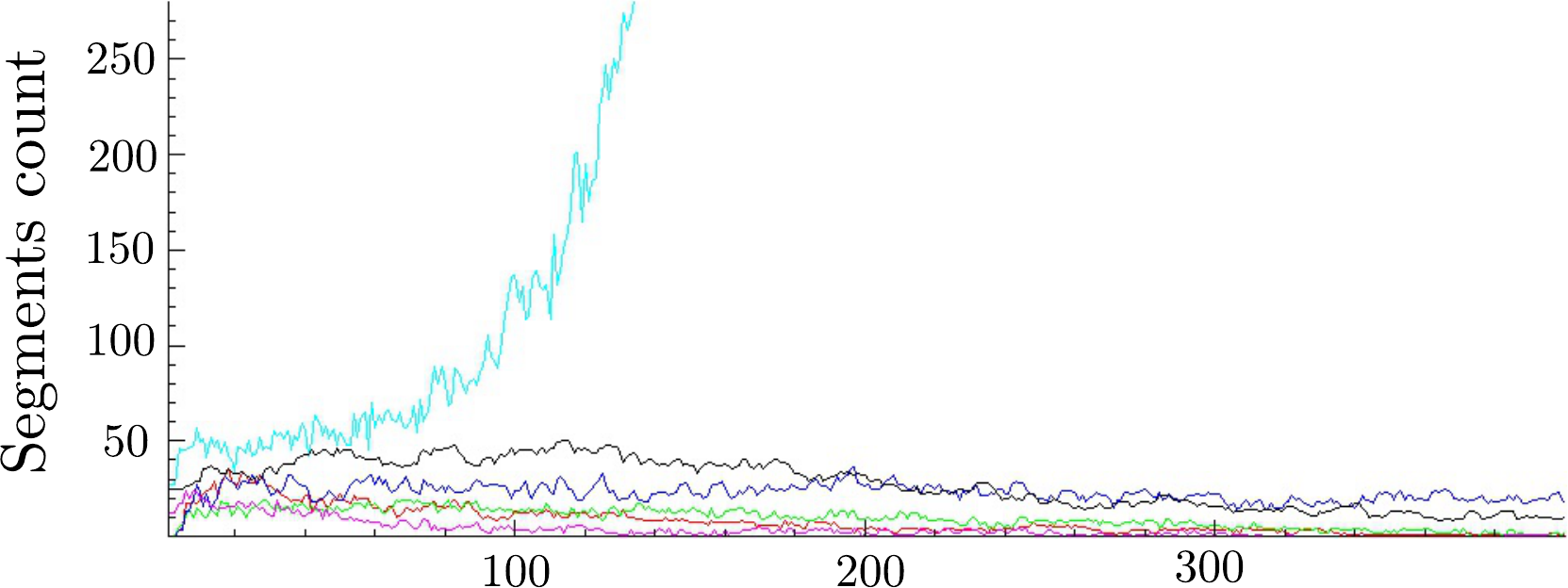} }
\subfigure[Number of split segments]{
\centering
  \includegraphics[width=0.68\linewidth]
      {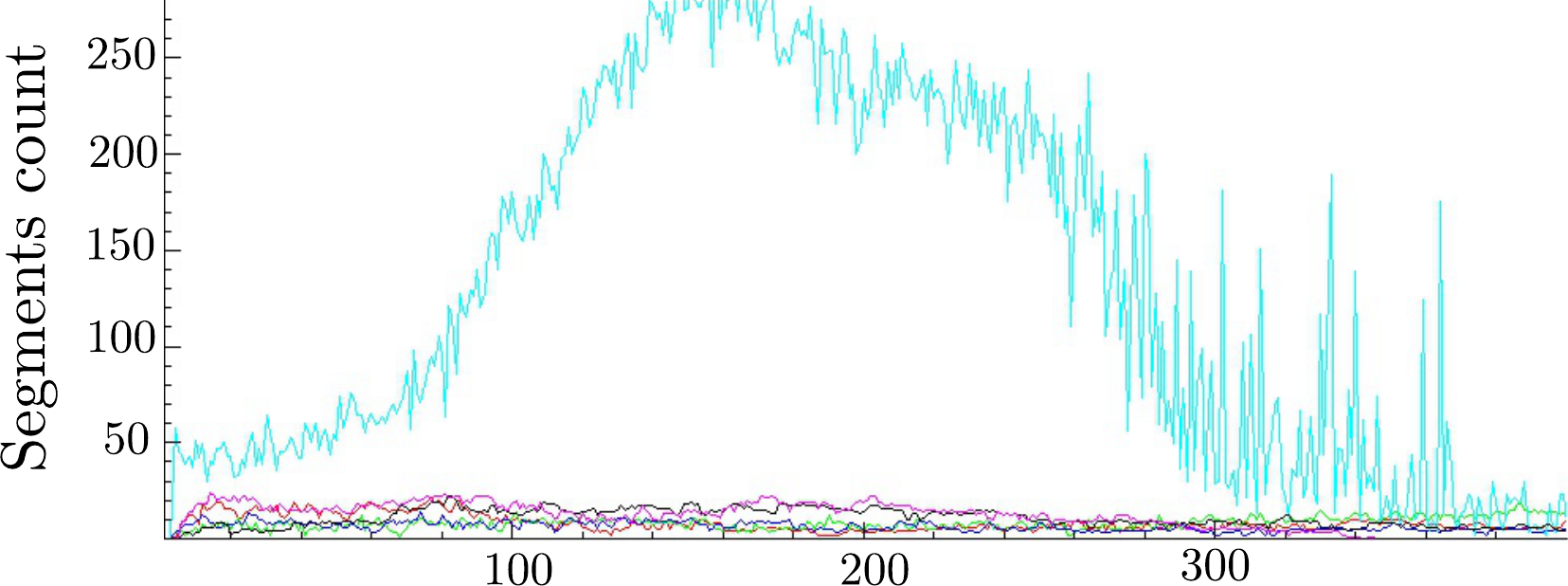} }
\centering
\caption{Comparison of the sensitivity to noise. The first data point is an
  image with already added noise, repeatability is computed using the results on
  the non noisy image given by each of the tested algorithm. The horizontal axis
  represents the amount of noise added to the image following
  equation~\ref{equ|sigmanoise}.}
\label{fig|seg|comp|noise}
\end{figure}

\paragraph*{Comparison.} Figure \ref{fig|seg|comp|noise} shows the comparison of
the sensitivity to the noise for the different algorithms. The number of
segments detected by \textit{Hough} increases drastically with the noise: this
produces a lot of unmatched segments as well as split segments. This is due to
the nature of the algorithm, since an increase of noise increases the number of
segment directions represented in the image -- this could however be alleviated
by adapting the parameters\footnote{The fact that the \textit{Hough} detector
  repeatability is better than all other detectors for indices greater than 150
  is an artifact caused by our similarity measure. This measure indeed does not
  favor long lines over small ones: when the noise increases, the \textit{Hough}
  detector generates numerous spurious small segments, among which the
  likelihood to have a segment matching the original is very high.}. The
\textit{LSD} algorithm is ``conservative'' to noise, which can be seen by the
rapid drop in number of detected segments, and therefore in repeatability, but
it has slightly less unmatched segments compared to
\textit{DSeg}. Unsurprisingly, the algorithms more robust with respect to noise
are \textit{DSeg} on images divided by four, and \textit{Hierarchical DSeg}. But
\textit{DSeg} on image divided by four yields a higher number of unmatched
segments, which might be explained by the decrease of the precision.

\subsubsection{Sensitivity to daylight changes}

\DONE{Indicate which image was used as reference}

In this experiment, images of a static scene have been acquired by a still
camera every 5 minutes over a lapse of 24 hours.

\begin{figure}[htb!]
\subfigure[Image 0]{
  \includegraphics[width=0.28\linewidth, trim=0 9mm 60mm 0]{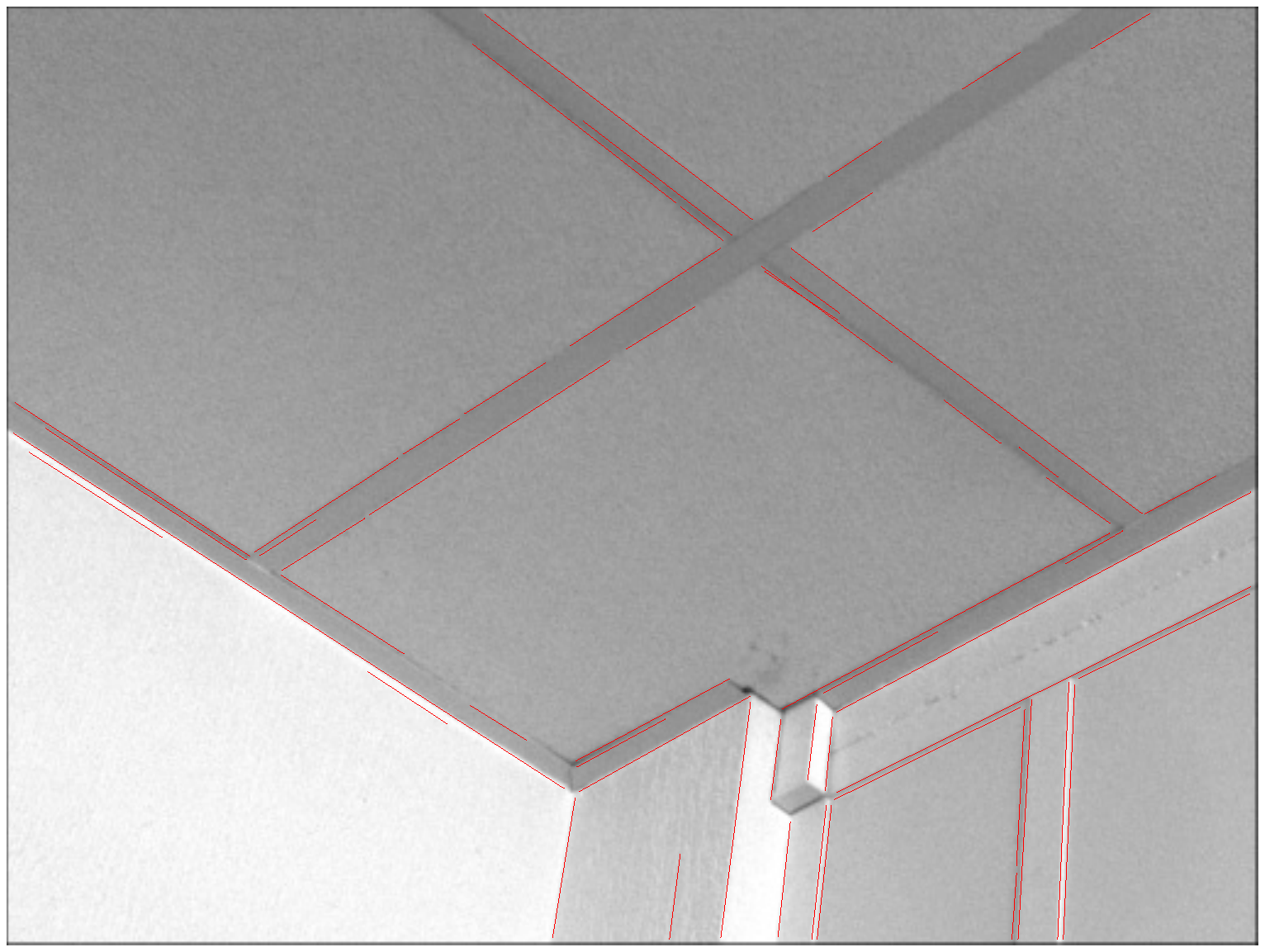} }
\subfigure[Image 14]{
  \includegraphics[width=0.28\linewidth, trim=0 9mm 60mm 0]{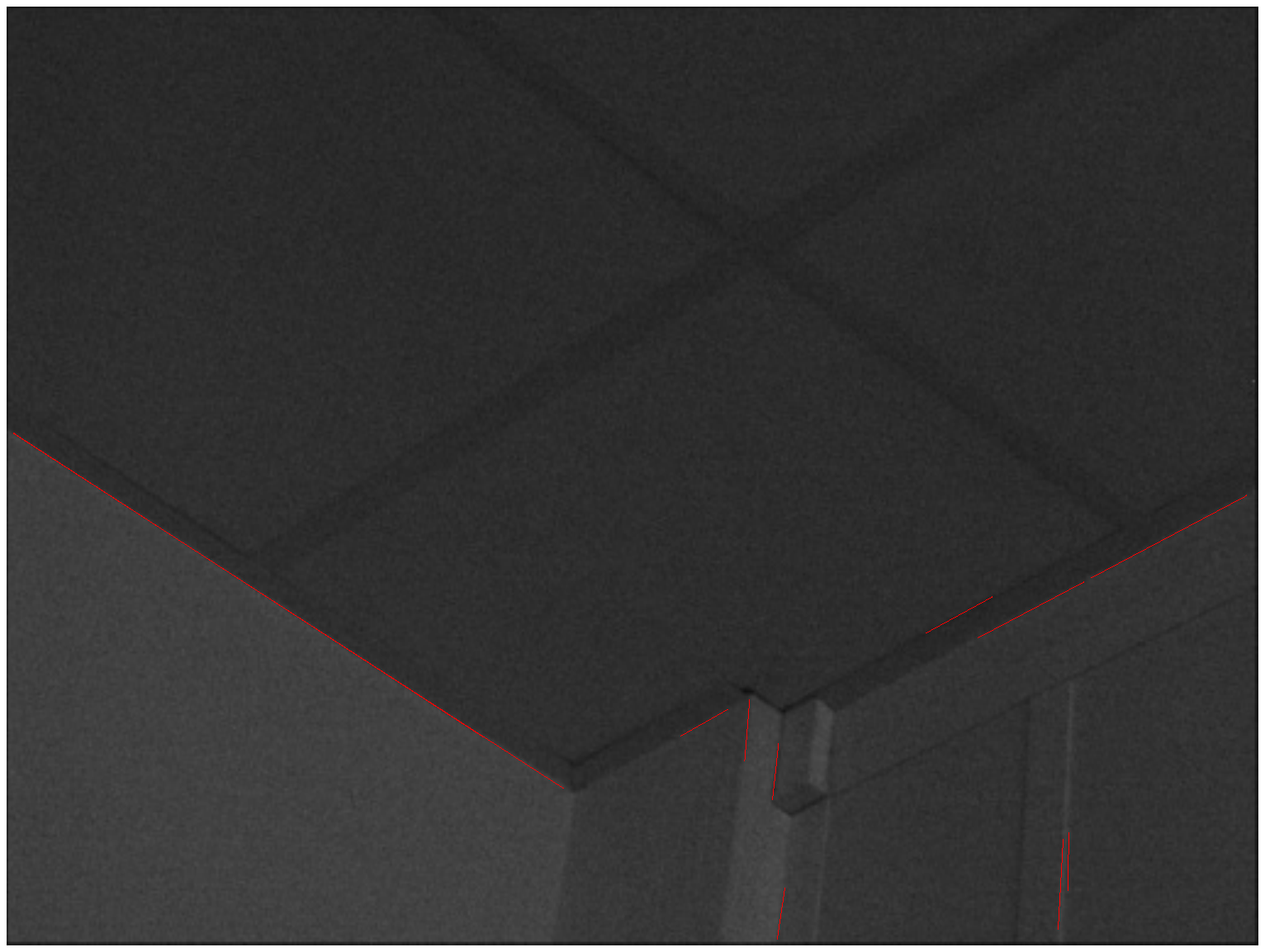} }
\subfigure[Image 200]{
  \includegraphics[width=0.28\linewidth, trim=0 9mm 60mm 0]{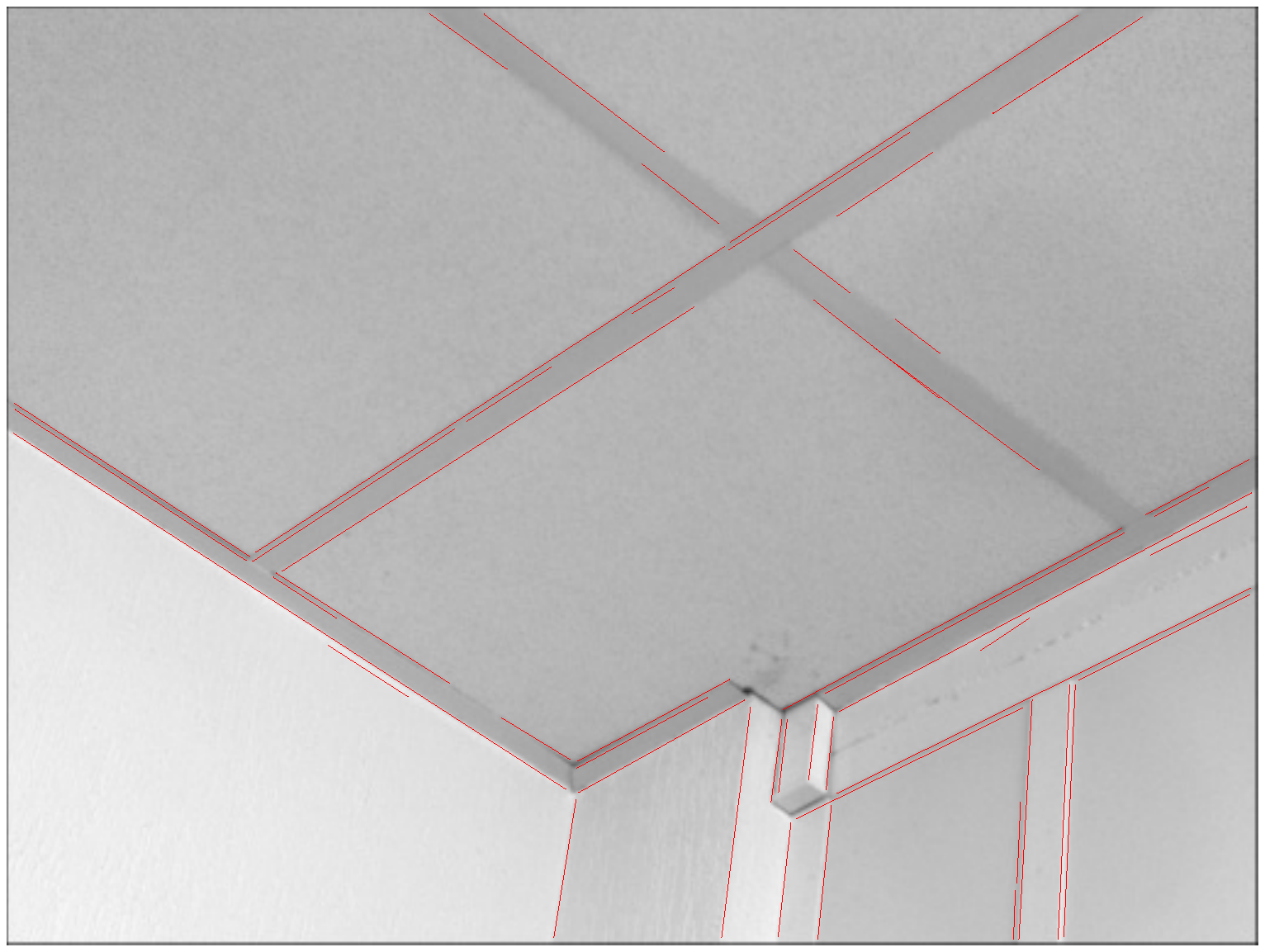} }
\centering \includegraphics[width=\linewidth, trim=0 195mm 50mm 0]{LegendSensitivity.pdf}
\subfigure[$ length = 20 $]{
  \centering
  \includegraphics[width=0.7\linewidth]
    {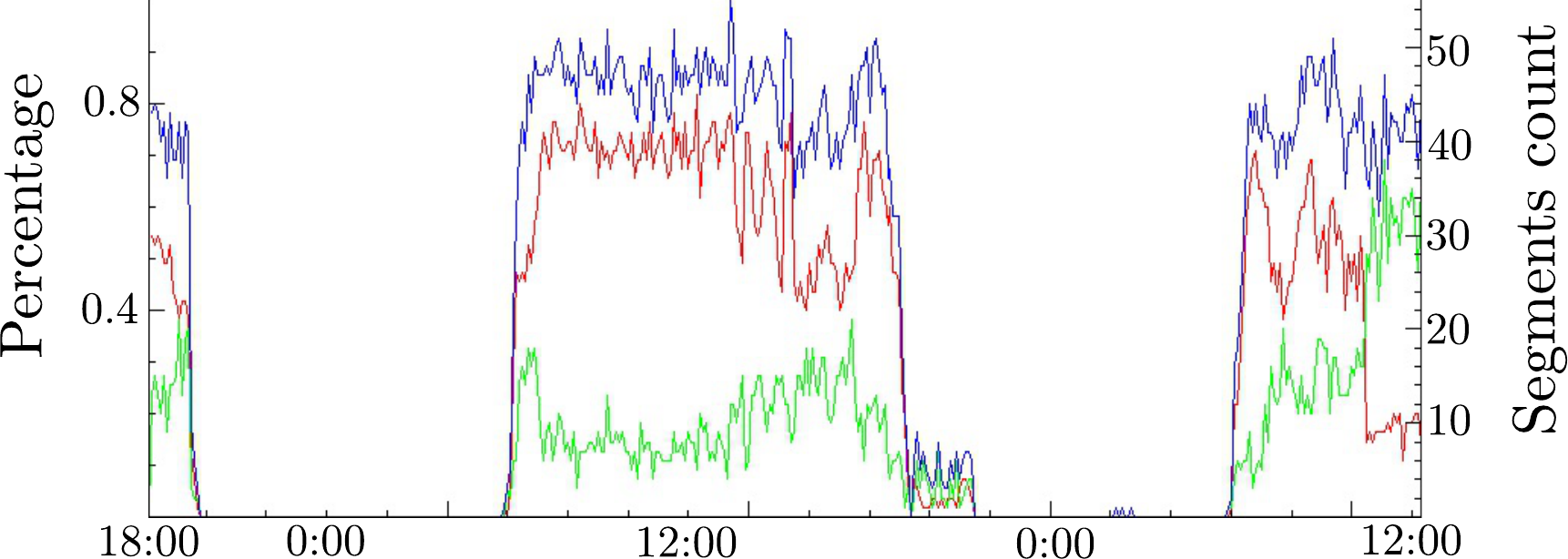} }
\subfigure[$ length = 50 $]{
  \centering
  \includegraphics[width=0.7\linewidth]
    {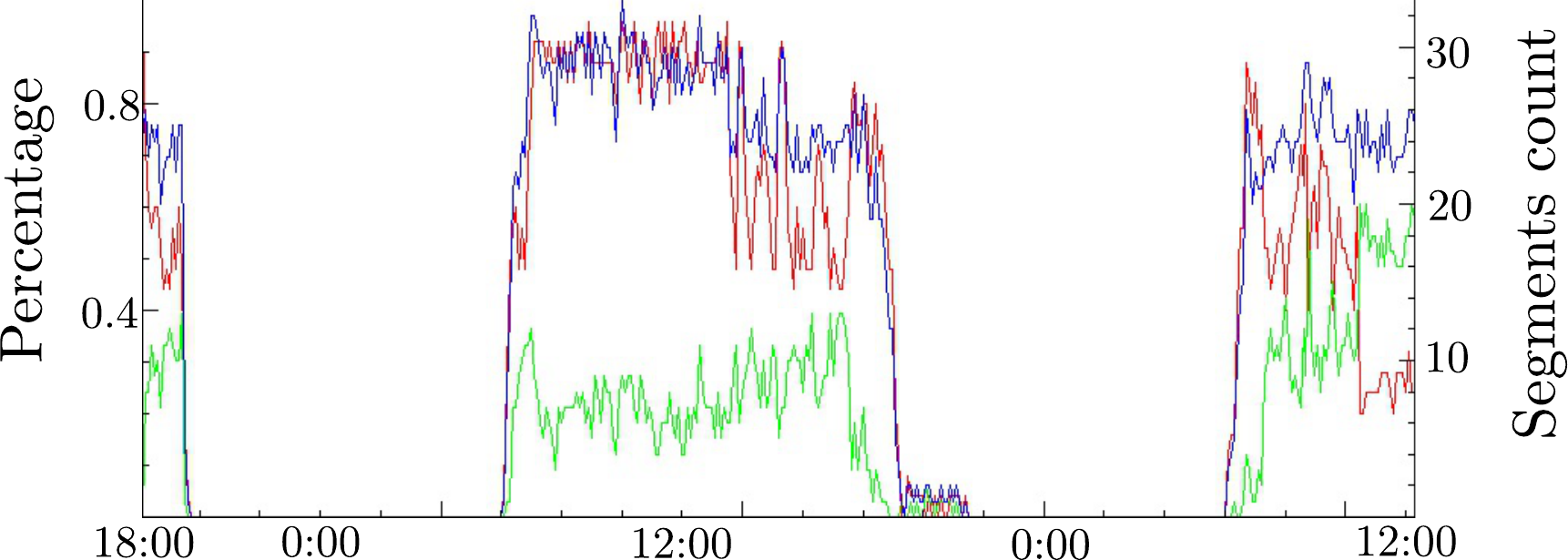} }
\subfigure[$ length = 100 $]{
  \centering
  \includegraphics[width=0.7\linewidth]
    {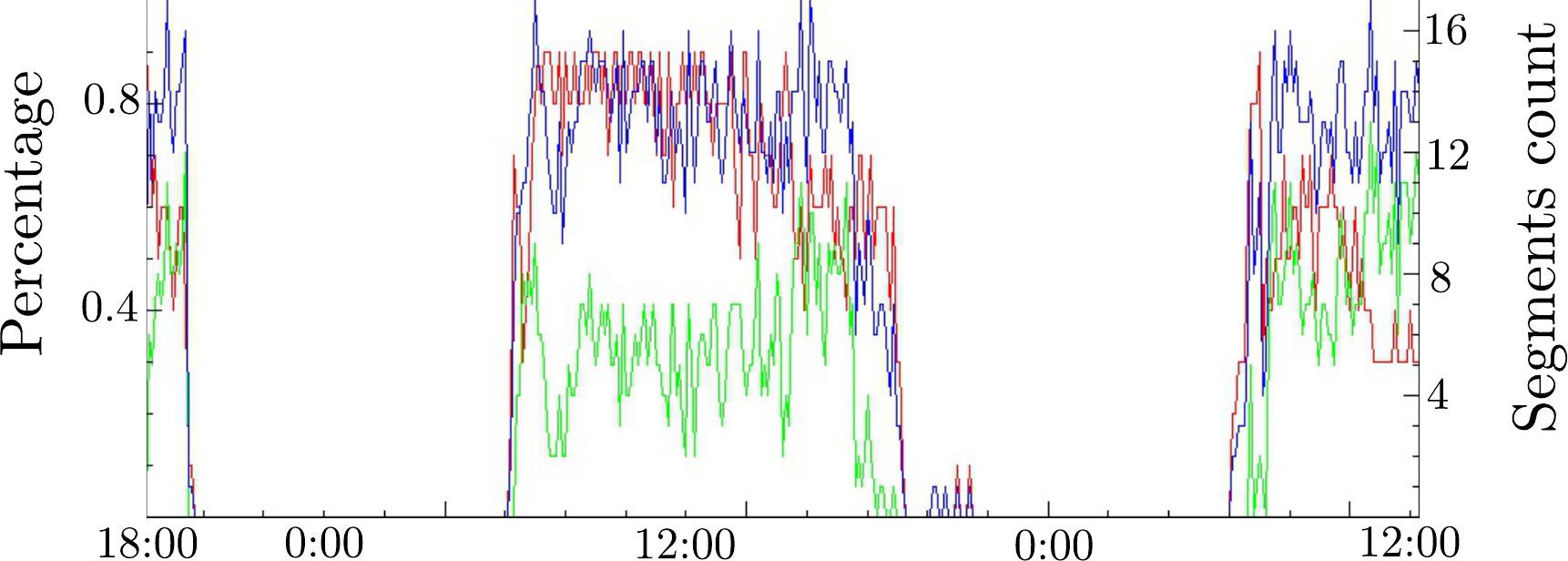} }
  \caption{Evolution of the repeatability of the detection by \textit{DSeg} with
    respect to illumination changes for different values of the minimal $length$
    of segments. The first image in the sequence is used as the reference for the repeatability.}
\label{fig|dseg|24|1}
\end{figure}

\begin{figure}[htb!]
\subfigure[Image 0]{
\includegraphics[width=0.28\linewidth, trim=0 9mm 60mm 0]
{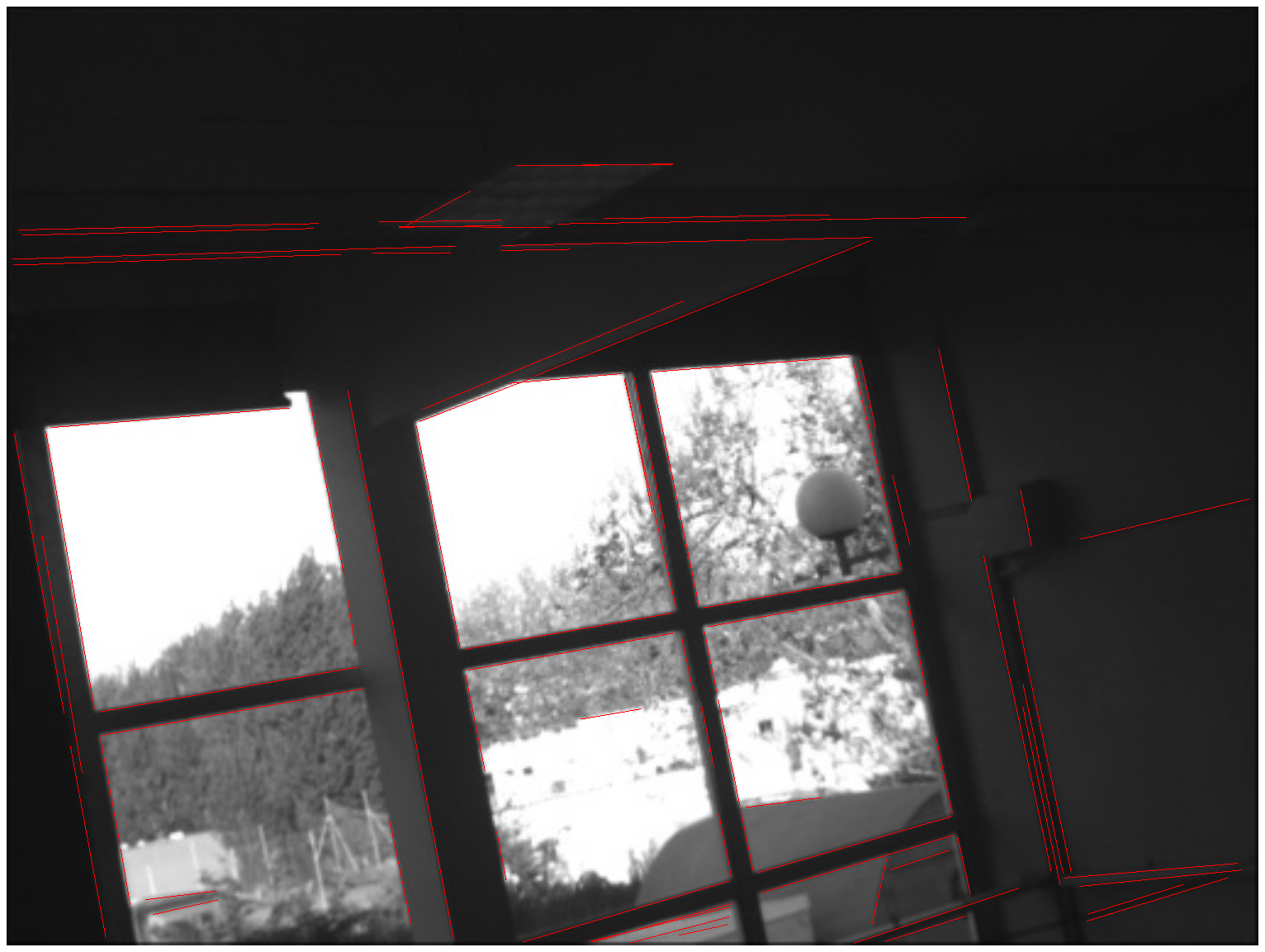} }
\subfigure[Image 100]{
\includegraphics[width=0.28\linewidth, trim=0 9mm 60mm 0]
{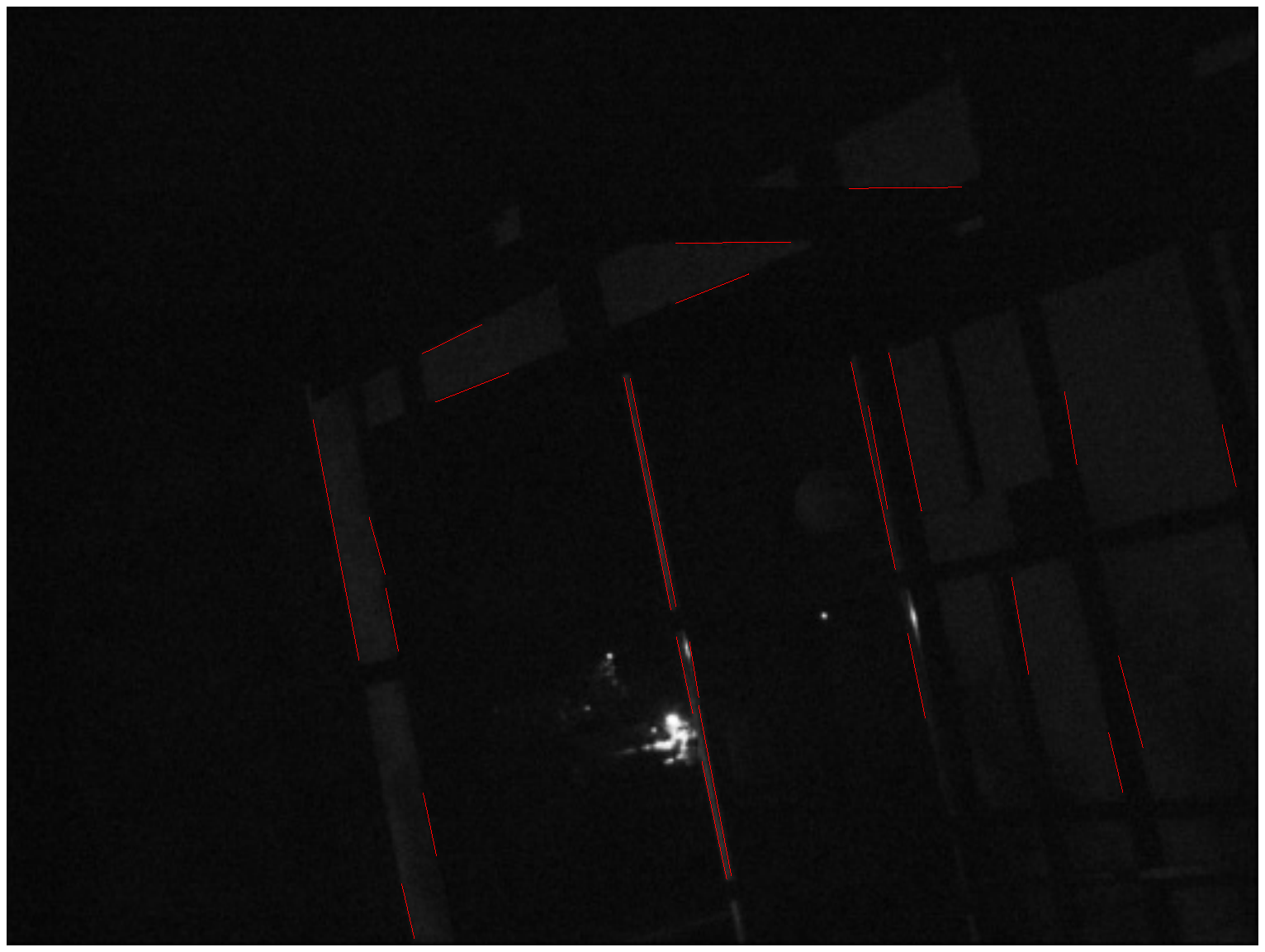} }
\subfigure[Image 200]{
\includegraphics[width=0.28\linewidth, trim=0 9mm 60mm 0]
{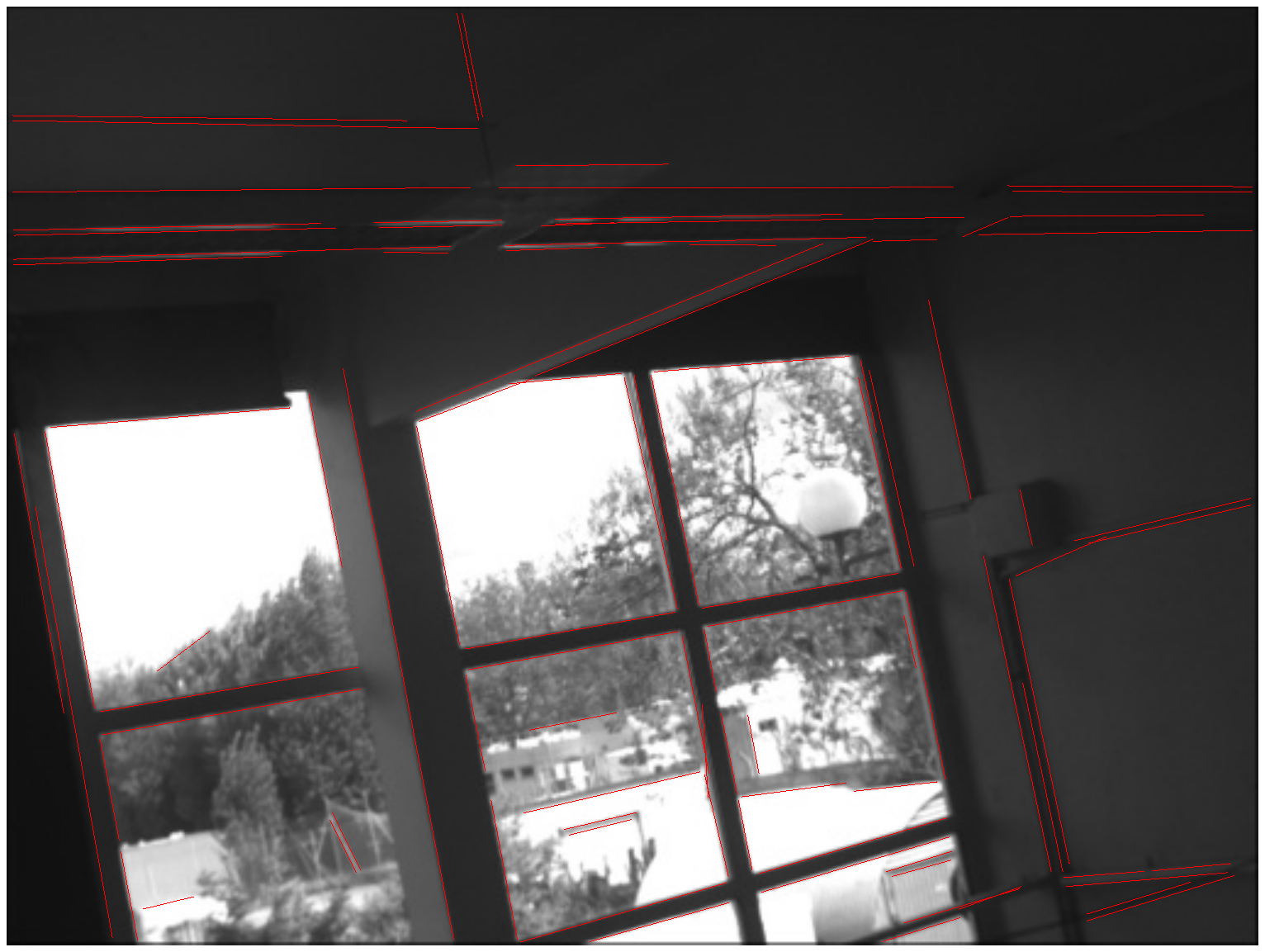} }
\centering \includegraphics[width=\linewidth, trim=0 195mm 50mm 0]{LegendSensitivity.pdf}
\subfigure[$ length = 20 $]{
  \centering
  \begin{tikzpicture}
    \begin{axis}[width=0.8\textwidth, height=4cm,
                legend columns=-1,
                legend to name=named_b,
                xmin = 0,
                xmax = 421,
                ymin = 0,
                ymax = 200,
                axis y line*=right,
                xtick={0, 50, 180, 310, 440},
                xticklabels={18:00, 0:00, 12.00, 0:00, 12.00},
                ylabel = Segments count]
      \addplot [color=blue, mark=none] table [x=a, y=b, col sep=space] {figures/results/24_comparison_length_b/detected.csv};
      \addlegendentry{Number of segments}
      \addplot [color=green, mark=none] table [x=a, y=b, col sep=space] {figures/results/24_comparison_length_b/unmatches.csv};
      \addlegendentry{Number of unmatched segments}
    \end{axis}
    \begin{axis}[width=0.8\textwidth, height=4cm,
                legend columns=-1,
                legend to name=named_b,
                xmin = 0,
                xmax = 421,
                ymin = 0,
                ymax = 1,
                axis y line*=left,
                xtick={0, 50, 180, 310, 440},
                xticklabels={18:00, 0:00, 12.00, 0:00, 12.00},
                ylabel = Percentage]
      \addplot [color=red, mark=none] table [x=a, y=b, col sep=space] {figures/results/24_comparison_length_b/repetability.csv};
      \addlegendentry{repetability}
    \end{axis}
  \end{tikzpicture}
  }
\subfigure[$ length = 50 $]{
  \centering
  \includegraphics[width=0.7\linewidth]{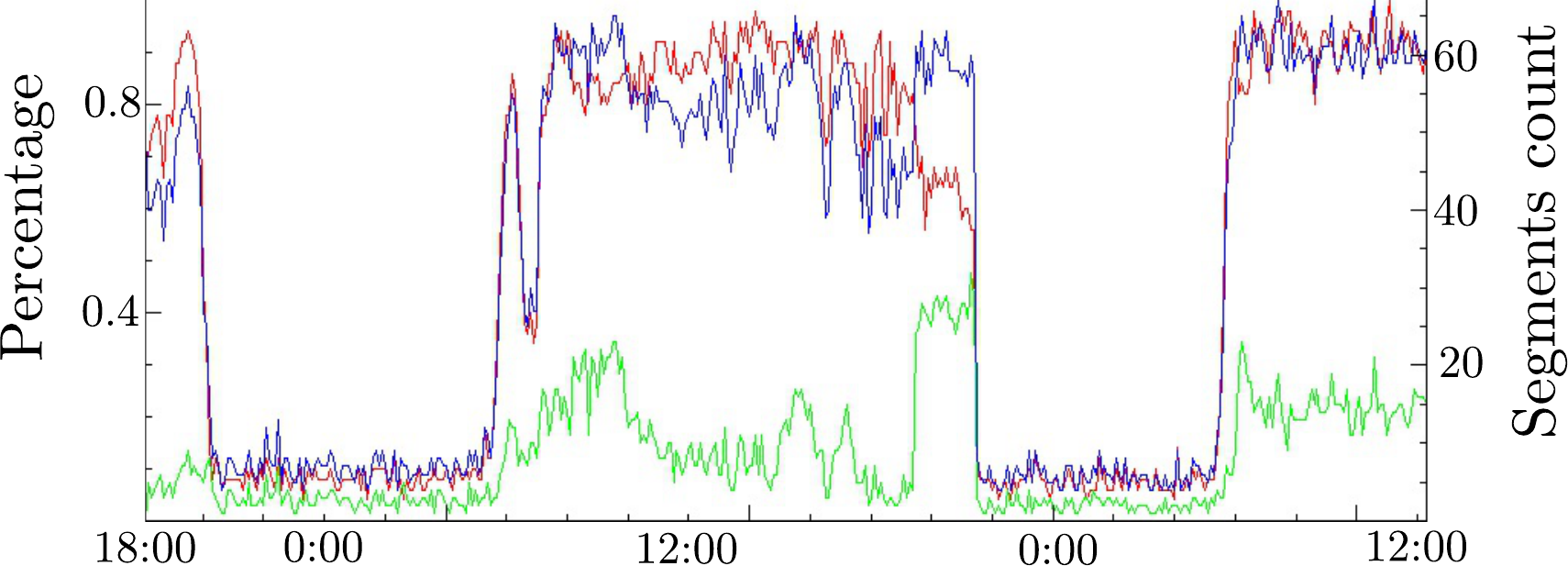}
  }
\subfigure[$ length = 100 $]{
  \centering
  \includegraphics[width=0.7\linewidth]{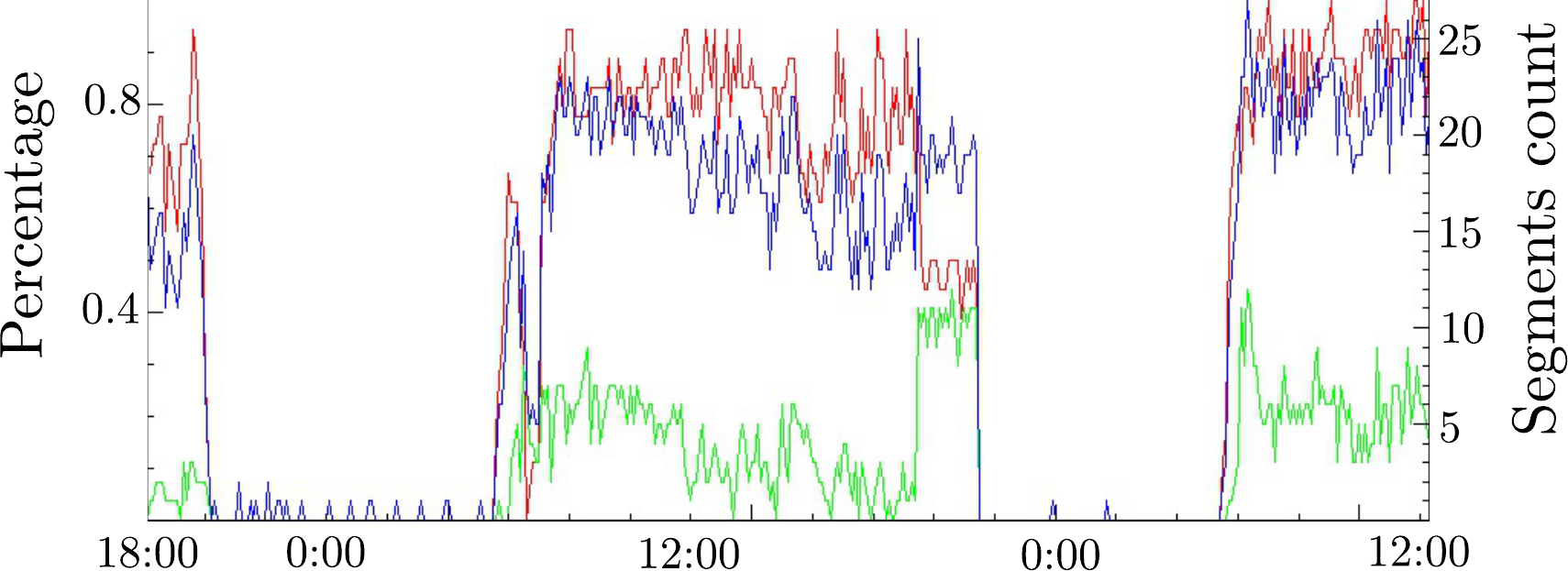}
  }
\ref{named_b}
\caption{Robustness to the change of light when changing the minimal $ length $ of segments with
\textit{DSeg}. The first image in the sequence is used as reference for repeatability.}
\label{fig|dseg|24|2}
\end{figure}

\paragraph*{Direct segment detection.} Figures \ref{fig|dseg|24|1} and
\ref{fig|dseg|24|2} show the evolution of the repeatability of \textit{DSeg}
over the 24 hours, for different minimal lengths of segments. On figure
\ref{fig|dseg|24|2} the drop visible at the end of the day is caused by the night
fall -- but since the lights were on, some segments (mostly inside the
room) are still detected. The figures show that the repeatability remains around
$0.9$ during daylight and that some segments are still correctly extracted under
very low light.

\paragraph*{Comparison.} Figure \ref{fig|seg|comp|daylight} shows how the
various algorithms perform with respect to illumination changes.
\textit{Hierarchical DSeg} detects as much segments as \textit{DSeg} on an image
divided by four, \textit{Hough} has the highest number of detected segments but
at the price of a lower repeatability and a higher number of split
segments. During the day \textit{Hierarchical DSeg}, \textit{DSeg} have a slighlty better repeatability than \textit{LSD} have a similar repeatability, which is higher than \textit{Chaining} and \textit{Hough}.
During the night, \textit{DSeg} on image divided by four yields the highest repeatability, followed by
\textit{Hierarchical DSeg}.

\begin{figure}[htb!]
\subfigure[Number of detected segments]{
\begin{tikzpicture}
  \begin{axis}[width=\textwidth, height=6cm,
               legend columns=-1,
               legend to name=named,
               xmin = 0,
               xmax = 287,
               ymin = 0,
               ymax = 200,
               xtick={0, 126 ,270},
               xticklabels={1:30, 12:00, 0.00},
               ylabel = Segments count]
    \addplot [color=red, mark=none] table [x=a, y=b, col sep=space] {figures/results/24_comparison/detected.csv};
    \addlegendentry{DSeg}
    \addplot [color=blue, mark=none ] table [ x=a, y=c, col sep=space] {figures/results/24_comparison/detected.csv};
    \addlegendentry{DSeg (scale 0.25)}
    \addplot [color=green, mark=none ] table [ x=a, y=d, col sep=space] {figures/results/24_comparison/detected.csv};
    \addlegendentry{HDSeg}
    \addplot [color=cyan, mark=none ] table [ x=a, y=e, col sep=space] {figures/results/24_comparison/detected.csv};
    \addlegendentry{Hough}
    \addplot [color=black, mark=none ] table [ x=a, y=f, col sep=space] {figures/results/24_comparison/detected.csv};
    \addlegendentry{Chaining}
    \addplot [color=gray, mark=none ] table [ x=a, y=g, col sep=space] {figures/results/24_comparison/detected.csv};
    \addlegendentry{LSD}
  \end{axis}
\end{tikzpicture}
}
\\[-2ex]
\subfigure[Repeatability]{
\begin{tikzpicture}
  \begin{axis}[width=\textwidth, height=4cm,
               xmin = 0,
               xmax = 287,
               ymin = 0,
               ymax = 1,
               xtick={0, 126 ,270},
               xticklabels={1:30, 12:00, 0.00},
               ytick={0, 0.25, 0.50, 0.75, 1.0},
               yticklabels={0\%, 25\%, 50\%, 75\%, 100\%},
               ylabel = Repeatability (\%)]
    \addplot [color=red, mark=none] table [x=a, y=b, col sep=space] {figures/results/24_comparison/repetability.csv};
    \addplot [color=blue, mark=none ] table [ x=a, y=c, col sep=space] {figures/results/24_comparison/repetability.csv};
    \addplot [color=green, mark=none ] table [ x=a, y=d, col sep=space] {figures/results/24_comparison/repetability.csv};
    \addplot [color=cyan, mark=none ] table [ x=a, y=e, col sep=space] {figures/results/24_comparison/repetability.csv};
    \addplot [color=black, mark=none ] table [ x=a, y=f, col sep=space] {figures/results/24_comparison/repetability.csv};
    \addplot [color=gray, mark=none ] table [ x=a, y=g, col sep=space] {figures/results/24_comparison/repetability.csv};
  \end{axis}
\end{tikzpicture}
}
\\[-2ex]
\subfigure[Number of unmatched segments]{
\begin{tikzpicture}
  \begin{axis}[width=\textwidth, height=3.2cm,
               xmin = 0,
               xmax = 287,
               ymin = 0,
               ymax = 80,
               xtick={0, 126 ,270},
               xticklabels={1:30, 12:00, 0.00},
               ylabel = Segments count]
    \addplot [color=red, mark=none] table [x=a, y=b, col sep=space] {figures/results/24_comparison/unmatches.csv};
    \addplot [color=blue, mark=none ] table [ x=a, y=c, col sep=space] {figures/results/24_comparison/unmatches.csv};
    \addplot [color=green, mark=none ] table [ x=a, y=d, col sep=space] {figures/results/24_comparison/unmatches.csv};
    \addplot [color=cyan, mark=none ] table [ x=a, y=e, col sep=space] {figures/results/24_comparison/unmatches.csv};
    \addplot [color=black, mark=none ] table [ x=a, y=f, col sep=space] {figures/results/24_comparison/unmatches.csv};
    \addplot [color=gray, mark=none ] table [ x=a, y=g, col sep=space] {figures/results/24_comparison/unmatches.csv};
  \end{axis}
\end{tikzpicture}
}
\\[-2ex]
\subfigure[Number of split segments]{
\begin{tikzpicture}
  \begin{axis}[width=\textwidth, height=3.2cm,
               xmin = 0,
               xmax = 287,
               ymin = 0,
               ymax = 80,
               xtick={0, 126 ,270},
               xticklabels={1:30, 12:00, 0.00},
               ylabel = Segments count]
    \addplot [color=red, mark=none] table [x=a, y=b, col sep=space] {figures/results/24_comparison/splited.csv};
    \addplot [color=blue, mark=none ] table [ x=a, y=c, col sep=space] {figures/results/24_comparison/splited.csv};
    \addplot [color=green, mark=none ] table [ x=a, y=d, col sep=space] {figures/results/24_comparison/splited.csv};
    \addplot [color=cyan, mark=none ] table [ x=a, y=e, col sep=space] {figures/results/24_comparison/splited.csv};
    \addplot [color=black, mark=none ] table [ x=a, y=f, col sep=space] {figures/results/24_comparison/splited.csv};
    \addplot [color=gray, mark=none ] table [ x=a, y=g, col sep=space] {figures/results/24_comparison/splited.csv};
  \end{axis}
\end{tikzpicture}
}
\\[-1ex]
\ref{named}
\caption{Comparison of the robustness to change of light (the reference image
is the 100th, where the repeatability is equal to one).}
\label{fig|seg|comp|daylight}
\end{figure}

\section{Conclusion}
We have proposed a new approach to extract line segments in images and an
extension to pyramidal detection.
The approach detects more and longer line segments than the classical
approaches, does not require any parameter tuning, and is robust with respect to
image noise and scene illumination variations. We have shown that our approach give
better results than state of the art algorithms.

The choice between \textit{Hierarchical DSeg} and \textit{DSeg} depends on whether the application
requires a larger number segments, or robust segments. Experiments show that
segment detection is as fast as classical interest point detectors. Most
importantly, segments exhibit more structure of the environment than points:
this is of great interest in numerous applications, such as scene interpretation
and mapping. For instance, our approach has been exploited in a vision-based
SLAM (Simultaneous Localisation and Mapping) approach \cite{SOLA-IROS-2009},
producing more useful maps than when using point features.

A demonstration of the algorithm is available at
\url{http://research.cberger.net/Line_Segments/Demo/}.

\bibliographystyle{elsart-num}
\bibliography{segments}

\end{document}